\documentclass{article} 
\usepackage{iclr2026_conference,times}

\usepackage{amsmath,amsfonts,bm}

\def\eqref#1{equation~\ref{#1}}

\def\1{\bm{1}}

\DeclareMathAlphabet{\mathsfit}{\encodingdefault}{\sfdefault}{m}{sl}
\SetMathAlphabet{\mathsfit}{bold}{\encodingdefault}{\sfdefault}{bx}{n}

\usepackage{hyperref}
\usepackage{url}

\usepackage{enumitem}
\usepackage{graphicx}
\usepackage{amsmath} 
\usepackage{multirow}
\usepackage{colortbl}
\usepackage{adjustbox} 
\usepackage{subcaption}
\usepackage{bbding}
\usepackage{pifont}
\usepackage{wrapfig}
\usepackage{amssymb} 
\usepackage{booktabs} 
\usepackage{algorithm}
\usepackage{algpseudocode}

\title{Personalized Vision via Visual In-context Learning}

\author{Yuxin Jiang$^{1,2}$
\quad
Yuchao Gu$^{1}$
\quad
Yiren Song$^{1}$
\quad
Ivor W. Tsang$^{2}$
\quad
Mike Zheng Shou$^{1\dagger}$\\\\
$^{1}$Show Lab, National University of Singapore
\quad
$^{2}$A*STAR, Singapore
}

\iclrfinalcopy 
\begin{document}

\maketitle

\vspace{-7mm}
\begin{figure}[ht]
  \centering
  \includegraphics[width=\linewidth]{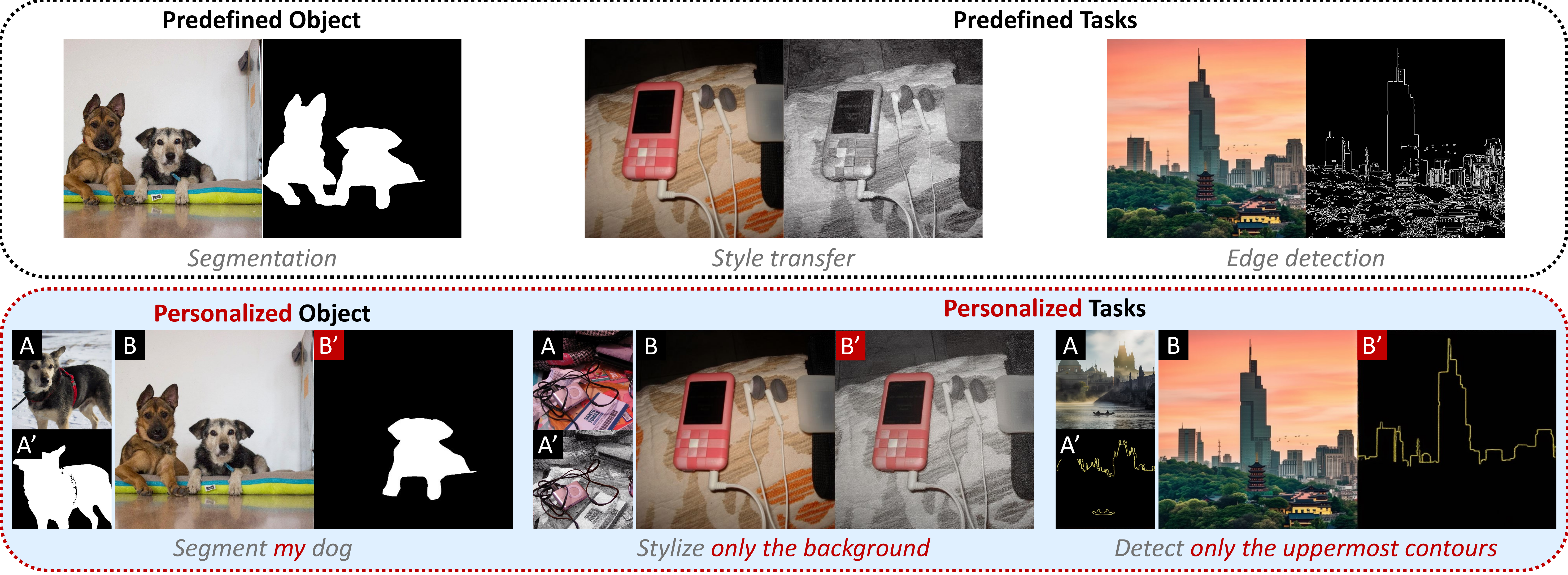}
  \caption{\textbf{Predefined vs. Personalized Vision.} Top: traditional, predefined tasks. Bottom: personalized tasks enabled by \textbf{PICO}. 
Given a new pair $(A \!\to\! A')$ and a query image $B$, our model \emph{infers the task in-context} and produces $B'$, adapting to novel user-defined tasks \emph{at test time}.
  }
  \label{fig:teaser}
\end{figure}

\begin{abstract}
Modern vision models, trained on large-scale annotated datasets, excel at predefined tasks but struggle with personalized vision—tasks defined at test time by users with customized objects or novel objectives.
Existing personalization approaches rely on costly fine-tuning or synthetic data pipelines, which are inflexible and restricted to fixed task formats. 
Visual in-context learning (ICL) offers a promising alternative, yet prior methods confine to narrow, in-domain tasks and fail to generalize to open-ended personalization.  
We introduce Personalized In-Context Operator (PICO), a simple four-panel framework that repurposes diffusion transformers as visual in-context learners. Given a single annotated exemplar, PICO infers the underlying transformation and applies it to new inputs without retraining. To enable this, we construct VisRel, a compact yet diverse tuning dataset, showing that task diversity, rather than scale, drives robust generalization.
We further propose an attention-guided seed scorer that improves reliability via efficient inference scaling.
Extensive experiments demonstrate that PICO (i) surpasses fine-tuning and synthetic-data baselines, (ii) flexibly adapts to novel user-defined tasks, and (iii) generalizes across both recognition and generation.
{Code: \url{https://github.com/showlab/PICO}}.
\end{abstract}

{\let\thefootnote\relax\footnotetext{$^\dagger$ Corresponding author.}}
\section{Introduction}
\label{intro}

Modern vision models~\citep{radford2021learning, oquab2024dinov2, kirillov2023segment, rombach2022high, esser2024scaling}, trained on large-scale annotated datasets, have achieved impressive performance in both visual recognition and generation. 
However, these models typically succeed on predefined object categories (\textit{e.g.}, cars, people) or standard task formats (\textit{e.g.}, object detection, semantic segmentation) where abundant labeled data exists.
They often struggle to adapt flexibly to \textbf{personalized vision\textemdash tasks defined by users at test-time}, involving customized objects or novel task definitions. 
With growing demand for personalized vision systems that quickly adapt to individual needs, a critical question emerges:
\textit{How can we achieve flexible and high-performing personalized vision?}

A traditional approach to personalized vision uses generative models to synthesize additional training data tailored to specific personalized objects. For example, Personalized Representation (PRPG)~\citep{sundaram2024personalized} employs DreamBooth~\citep{ruiz2023dreambooth} to generate synthetic data for target concepts, then adapting general-purpose feature representations into personalized ones.
While these methods~\citep{sundaram2024personalized, zhang2024low} make strides toward personalized vision by adapting to personalized objects, they remain constrained to predefined task (\textit{e.g.}, segmentation or classification) and require costly fine-tuning for each new subject.
They do not generalize flexibly to arbitrary, user-defined tasks.

In natural language processing (NLP), in-context learning (ICL)~\citep{brown2020language, dong2024survey} has shifted practice from task-specific fine-tuning toward models that can perform novel tasks defined at test time.
A natural analogy in vision is to let exemplars define the task.  However, unlike text, vision tasks have heterogeneous output format (\textit{e.g.,} pixel arrays, masks, coordinates), making in-context generalization more challenging. 
Existing visual ICL methods~\citep{bar2022visual, wang2023images, bai2024sequential} unify tasks but fall short of personalized vision:
they are typically evaluated on predefined, narrow, in-domain tasks and show limited generalization beyond training set, rather than adapting to open-ended personalized tasks at test time.
%

To address this gap, we study personalized task generalization: adapting to novel tasks or novel objects during test time. 
We introduce the \textbf{Personalized In-context Operator (PICO)}, a simple visual ICL framework based on a four-panel input format, where an annotated exemplar $(A \!\to\! A')$ defines the task and the model infers the underlying transformation and applies it to new inputs $(B \!\to\! B')$. 
To support this setting, we construct the \textbf{VisRel dataset}, a compact yet diverse tuning dataset of structurally organized visual tasks, designed to expose the model to a unified \emph{visual-relation space} for broad generalization. 
To mitigate stochastic sampling variability, we propose an \textbf{attention-guided seed scorer} that leverages early-step cross-grid attention patterns to rank candidate seeds, enabling efficient test-time scaling.

We conduct extensive experiments to validate the effectiveness of PICO. 
First, PICO outperforms fine-tuning–based approaches on personalized subjects within conventional vision tasks. 
Second, PICO flexibly adapts to novel, user-defined tasks at test time, supported by both quantitative and qualitative results. 
Finally, PICO achieves strong performance across diverse personalized vision scenarios, covering both recognition and generation.

In summary, our key contributions are:
\begin{itemize}
    \item We formulate \textbf{personalized vision as visual in-context learning}, enabling a single generative prior to adapt at test time to both new objects and new tasks from exemplars, without requiring synthetic data or costly per-subject fine-tuning.
    \item We construct the \textbf{VisRel dataset}, a compact yet diverse tuning dataset, and show that task diversity, rather than scale, drives strong generalization in visual ICL. 
    \item We propose an \textbf{attention-guided seed scorer} that leverages early attention dynamics, improving the reliability stochastic generative sampling through efficient inference scaling.
    \item We demonstrate, across diverse benchmarks, that PICO achieves strong personalization performance with minimal supervision, spanning both recognition and generation, and covering varied subjects and task definitions. 
\end{itemize}
\section{Related Work}
\label{related}

\noindent{\bf  Personalized Vision.}
Existing personalized vision methods~\citep{sundaram2024personalized, zhang2024low, alaluf2024myvlm, cohen2022my, nguyenyo, zhangpersonalize, samuel2024s} typically adapt vision or vision-language models (VLMs) to handle user-specific concepts within predefined tasks like retrieval and segmentation.
For example, PerSAM~\citep{zhangpersonalize} segments user-indicated regions using cosine similarity on pretrained segmentation features~\citep{kirillov2023segment}, while PDM~\citep{samuel2024s} leverages intermediate features from text-to-image (T2I) models~\citep{rombach2022high} to localize personalized instances.
PRPG~\citep{sundaram2024personalized} generates synthetic training data to enhance personalized representations for downstream tasks.
However, these methods are inherently restricted to fixed task formats, lacking flexibility to accommodate arbitrary user-defined tasks at test-time.
Real-world personalization often demands versatile, dynamically defined tasks (\textit{e.g.}, inserting custom objects, generating annotations in new formats).  
Such scenarios motivate our approach to enable personalized vision systems to rapidly adapt beyond fixed frameworks.

\noindent{\bf Visual In-Context Learning.}  Visual ICL, inspired by prompt-based task adaptation in NLP~\citep{brown2020language}, aims to adapt vision models to downstream tasks through contextual examples.
Bar \textit{et al.}~\citep{bar2022visual} first propose visual prompting by framing vision tasks as quad-grid masked image inpainting.
Painter~\citep{wang2023images}, a ViT-based model~\citep{dosovitskiy2020vit} trained through masked image modeling, shows strong ICL capabilities across various dense prediction tasks, and SegGPT~\citep{SegGPT} further enhances this ability specifically for segmentation.
However, these training-based visual ICL methods ~\citep{bar2022visual, wang2023images, SegGPT} rely heavily on extensive, task-specific pretraining, limiting generalization to unseen tasks. 
In contrast, inference-based methods~\citep{nguyen2023visual, yang2023imagebrush, zhao2024instructbrush, gu2024analogist, lai2024unleashing} attempt to interpret visual demonstrations by translating them into textual instructions, which underuses visual signals and remains confined to semantic
editing tasks, leading to inaccuracies from ambiguous text descriptions.
%
%
Our work advances visual ICL by explicitly formulating personalized vision as visual relations within a unified space, enabling robust, flexible one-shot personalization tailored to individual needs.

\noindent{\bf Diffusion Priors.} Diffusion models have emerged as the defacto paradigm for image synthesis~\citep{rombach2022high, esser2024scaling}, demonstrating powerful generative priors beneficial for diverse vision tasks, including dense prediction~\citep{he2024lotus, fu2024geowizard, ke2023repurposing}, image restoration~\citep{xia2023diffir, wang2024sinsr, he2025diffusion, lugmayr2022repaint}, style transfer~\citep{chung2024style, jiang2025balanced, song2025omniconsistency}, etc.
Within data-scarce personalized vision settings, diffusion models are commonly employed to synthesize additional training data for downstream fine-tuning~\citep{ruiz2023dreambooth, galimage}.
However, this two-stage process~\citep{sundaram2024personalized} is computationally intensive, limiting practicality for frequent adaptation to personalized concepts.
Recent work such as In-Context LoRA~\citep{huang2024context, zhang2025easycontrol} have highlighted intrinsic ICL ability in diffusion transformers~\citep{peebles2023scalable}. 
Building on these insights, we directly use diffusion priors as visual in-context learners, enabling flexible, immediate adaptation to arbitrary user-defined tasks without synthetic augmentation or retraining.

\begin{figure}[!b]
  \centering
  \includegraphics[width=\linewidth]{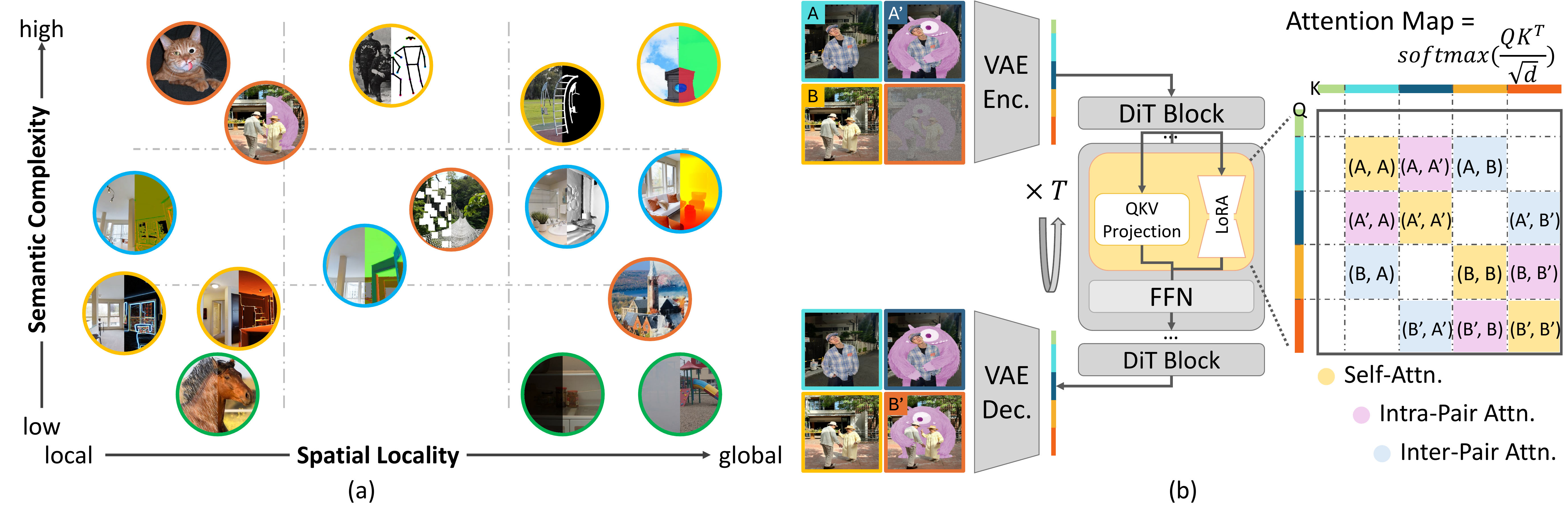}
  \vspace{-5mm}
  \caption{
  \textbf{(a) Structured Visual Relation Space.} Tasks are organized by semantic complexity (low to high) and spatial locality (local to global), covering diverse task types, color-coded as:
  \textcolor{green}{\ding{110}} restoration/enhancement, 
  \textcolor{cyan}{\ding{110}} physical/geometric estimation, 
  \textcolor{yellow}{\ding{110}} semantic perception, 
  \textcolor{orange}{\ding{110}} generative manipulation. 
  \textbf{(b) Training pipeline of PICO.}}
  \label{fig:main}
\end{figure}
\vspace{-3mm}
\section{Method}
\label{method}
Our objective is to achieve flexible visual personalization through a task-agnostic framework that adapts to user-defined tasks at inference, without additional fine-tuning.
We reformulate personalized vision as a visual ICL problem, where a single input-output exemplar defines the task, and the model infers user intent from this demonstration and applies it to new queries.
Central to our approach is learning a broad \textit{visual-relation space}, repurposing pretrained diffusion transformers into in-context visual reasoners. 
We further introduce a lightweight seed-selection strategy for inference scaling that enhances stability and reliability. 

\subsection{Data: A Visual Relation Space}
ICL succeeds in NLP because every task (\textit{e.g.}, translation, summarization, question answering, etc.) shares a unified language generation interface.
In vision, however, different tasks have heterogeneous output format (\textit{e.g.}, pixel arrays, masks, coordinates), limiting the potential for unified in-context generalization.
We address this by unifying visual tasks as image-to-image transformations represented as RGB inputs and outputs~\citep{bar2022visual, wang2023images}.
Our key insight is that a robust visual ICL model should similarly embed tasks within a unified visual relation space, enabling interpolation and composition of transformations at test time.
To learn this space, we curate VisRel, a compact yet diverse dataset of 27 visual tasks, aiming to span the space of common 2D transformations (see Figure~\ref{fig:main}(a)). Its design follows three principles.
%
%

\noindent{\bf{Task Taxonomy.}} We structure the visual relation space along two intuitive axes: 
(1)~\textit{Semantic Complexity} measures the level of semantic understanding required, spanning low-level (pixel/color adjustments), mid-level (structure/shape manipulation), to high-level (object/class reasoning) transformations. 
(2)~\textit{Spatial Locality} defines the spatial context dependency, ranging from local (neighboring pixels), intermediate (objects patches), to global (full-image context) operations.

\noindent{\bf{Intra-task Diversity.}} 
Each task includes diverse variants to avoid overfitting.
For instance, inpainting uses masks of varying colors, shapes, and transparency; 
segmentation supports different colors, transparency mask; 
restoration tasks (denoising, deblurring) include multiple noise levels or blur kernels. 
%
This encourages learning transferable transformation principles rather than memorizing task-specific patterns, which is important for zero-shot generalization to novel personalized tasks.

\noindent{\bf{Minimal Text Label.}} 
The model primarily relies on visual exemplars, but minimal text prompts help resolve ambiguities between potential conflicts of interest tasks (\textit{e.g.}, local vs. global edits; black and white depth estimation vs.colorful style transfer).
These lightweight cues (\textit{e.g.}, “edit” vs. “estimate”) act as soft boundaries while keeping the framework largely vision-driven.

%
%

\subsection{Training: PICO}
\label{sec:training}

Given an exemplar pair $\{A, A'\}$ illustrating a visual relation $r: A \!\to\! A'$ and a query image $B$, the goal is to synthesize an output $B'$ that applies $r$ to $B$.
We adopt a quad-grid input format
\begin{equation}
\label{eq:grid}
I = \textsc{Grid}\!\left(\begin{bmatrix} A & A' \\ B & B' \end{bmatrix}\right).
\end{equation}
The training pipeline is illustrated in Figure~\ref{fig:main}(b). We build upon a pretrained  T2I diffusion transformer (DiT)~\citep{flux2024}, finetuned with LoRA~\citep{hu2022lora}.
A VAE encoder $\mathcal{E}(\cdot)$ maps the grid into latent space, yielding the target $x_0=\mathcal{E}(B')$ and visual conditions $c_{\text{vp}}=\mathcal{E}(\{A,A',B\})$, while $c_\text{txt}$ encodes minimal text prompts.
At time $t$, the latent sequence is $Z_t = [\,x_t;\; c_{\text{vp}};\; c_\text{txt}\,]$, where the target latent is noised as $x_t = (1-t)x_0 + t\,\epsilon$, with $\epsilon \sim \mathcal{N}(0,I)$ and $ t \sim \mathcal{U}(0,1)$.
 Each DiT block applies multi-modal attention. For head $h$ in block $b$,
\begin{equation}
\label{eq:mma}
\operatorname{MMA}^{(b,h)}(Z_t) = \mathrm{softmax}\!\Big(\tfrac{Q^{(b,h)}_t {K^{(b,h)}_t}^\top}{\sqrt{d_h}}\Big) V^{(b,h)}_t,
\end{equation}
where $Q_t,K_t,V_t$ are projections of $Z_t$, $H$ is the number of heads.

\noindent{\textbf{Clean noising and Objective.}}
Unlike In-Context LoRA~\citep{huang2024context}, which perturbs all latents, we inject noise solely into the target $x_0$, leaving $c_{\text{vp}},c_\text{txt}$ clean. This prevents corruption of exemplar information and and yields stable relation transfer. 
The training objective is applied only on the target quadrant, so the model focuses on reconstructing $B'$ while leveraging the clean context for guidance. Concretely, the model predicts a conditional velocity field $v = v_\Theta(x_t, t \mid c_\text{txt}, c_{\text{vp}}),$, trained with conditional flow matching (CFM):
\begin{equation}
\label{eq:loss_cfm}
\mathcal{L}_{\text{CFM}} = \mathbb{E}_{t, x_t}\!\left[\| v_\Theta(x_t, t \mid c_\text{txt}, c_{\text{vp}}) - \hat v(x_t, t)\|^2\right],
\end{equation}
where $\hat v(\cdot)$ is the oracle velocity defined by the flow schedule.

\subsection{Inference: One-shot personalization.}

At test time, the $B'$ quadrant is replaced by a placeholder $X$, initialized as Gaussian noise in latent space. The three context quadrants remain clean: $c_{\mathrm{\text{vp}}} = \mathcal{E}([A, A', B])$. Starting from $x_1\!\sim\!\mathcal{N}(0,I)$, we integrate the learned flow
\begin{equation}
\label{eq:flow}
\frac{dx_t}{dt}=v_\Theta(x_t,t\mid c_\text{txt},c_{\mathrm{\text{vp}}}),
\end{equation}
from $t=1\!\to\!0$ to obtain $x_0$, and decode $B'=\mathcal{D}(x_0)$, where $\mathcal{D}$ is the VAE decoder.  The model seamlessly transfers the visual transformation demonstrated by $(A, A')$ to the query $B$, supporting flexible, test-time personalization without fine-tuning.

\noindent\textbf{{Inference scaling via Attention-Guided Seed Selection.}}
Generative sampling is stochastic: different seeds can diverge, which is undesirable for deterministic or localized tasks (see Figure~\ref{fig:abl_tts}).
We introduce a training-free seed scorer that exploits early cross-attention routing (Eq.~\ref{eq:mma}) to select promising seeds before full denoising. 
We refer to the bottom-right (BR) quadrant as the target region.
During training, BR contains the ground-truth $B'$; during inference, BR is the placeholder $X$.
Our intuition is that BR queries should initially bind to evidence in $B$ and transformation cues in $(A,A')$, then pivot back to BR; persistent focus on exemplars risks copying rather than adapting.

Let $p^{\text{br}}_{s,b,i}$ and $p^{\text{\text{vp}}}_{s,b,i}$ denote the average attention mass from target BR queries to BR keys and to visual context keys ($A,A',B$), respectively, at early solver step $i$ for seed $s$ (averaged over heads $H$). 
For blocks $\mathcal{B}^{\dagger}$ and the first few solver steps $i\in\{0,1,2\}$, we measure the pivot:
\begin{equation}
\label{eq:grow-D}
D_{\text{br}}(s) = \tfrac{1}{|\mathcal{B}^\dagger|}\sum_{b\in\mathcal{B}^\dagger}\!\big(p^{\text{br}}_{s,b,2}-p^{\text{br}}_{s,b,0}\big), \quad
D_{\text{vp}}(s) = \tfrac{1}{|\mathcal{B}^\dagger|}\sum_{b\in\mathcal{B}^\dagger}\!\big(p^{\text{vp}}_{s,b,2}-p^{\text{vp}}_{s,b,0}\big).
\end{equation}
Here, $D_{\text{br}}(s)$ quantifies how strongly queries pivot toward the target, while $D_{\text{vp}}(s)$ captures how quickly they peel away from exemplars. The final seed score is
\begin{equation}
\label{eq:pivot}
S_{\mathrm{pivot}}(s) = z(D_{\text{br}}(s)) - z(D_{\text{vp}}(s)), \quad
s^\star = \arg\max_{s\in\mathcal{S}} S_{\mathrm{pivot}}(s),
\end{equation}
with $z(\cdot)$ denoting $z$-normalization across candidate seeds $\mathcal{S}$. Pseudo-code and further statistical analysis are in the Appendix~\ref{app:tts}.
\section{Experiments}
\label{exp}

We validate our method through extensive experiments addressing three key questions:
(1) Does visual ICL surpass traditional personalized fine-tuning on standard tasks like personalized segmentation?
(2) Can the framework handle novel, user-defined tasks at inference?
(3) Does it extend across recognition and generation tasks?

\subsection{Implementation Details}
\label{exp_details}
We build PICO upon FLUX.1-dev~\citep{flux2024}, a latent rectified flow transformer model, finetuning with LoRA~\citep{hu2022lora} (rank 256) on the VisRel dataset for $30,000$ steps using a single H100 GPU.
All experiments are conducted at a resolution of $1024 \times 1024$, with each cell of the quad-grid structured as $512 \times 512$.
We use the Prodigy optimizer~\citep{mishchenko2024prodigy} with safeguard warmup, bias correction enabled, and a weight decay of $0.01$.
The VisRel training dataset contains $315$ samples across $27$ diverse tasks, curated from existing sources. Details of data construction are provided in Appendix~\ref{visrel}.
%
For fair comparison, we report results using a single default seed. Results annotated with \textsc{TTS} additionally employ our proposed test-time scaling strategy. In this setting, we fix $\mathcal{B}^\dagger=\{9,11,12\}$, probe steps $\{0,1,2\}$, and use a candidate seed set of size $|\mathcal{S}|=10$.  
Code and model will be released.

\newcommand{\deltaup}[1]{\textcolor{red}{\small$+$\,#1}}

\begin{table}[!t]
  \caption{\textbf{Quantitative Comparison on personalized segmentation.} We compare PICO with diverse baselines.
  \ding{72}: best, \ding{73}: second-best, and  \ding{117}: third-best.}
  \vspace{-3mm}
  \centering
  \begin{adjustbox}{width=\linewidth}
  \begin{tabular}{@{}lcccccccccccc@{}}
    \toprule
    \multirow{2}{*}{Method} & \multicolumn{3}{c}{PerSeg} & \multicolumn{3}{c}{DOGS} & \multicolumn{3}{c}{PODS} & \multicolumn{3}{c}{PerMIS} \\
    \cmidrule(lr){2-4} \cmidrule(lr){5-7} \cmidrule(lr){8-10} \cmidrule(lr){11-13}
     & mIOU$\uparrow$ & bIOU$\uparrow$ & F1$\uparrow$ & mIOU$\uparrow$ & bIOU$\uparrow$ & F1$\uparrow$ & mIOU$\uparrow$ & bIOU$\uparrow$ & F1$\uparrow$ & mIOU$\uparrow$ & bIOU$\uparrow$ & F1$\uparrow$ \\
    \midrule
    \textit{large-scale} \\
    PerSAM & 90.50\textsuperscript{\ding{117}} & 72.79\textsuperscript{\ding{117}} & 94.07\textsuperscript{\ding{73}} & 86.87\textsuperscript{\ding{73}} & 71.06\textsuperscript{\ding{72}} & 53.18 & 67.45\textsuperscript{\ding{73}} & 56.63\textsuperscript{\ding{73}} & 45.60\textsuperscript{\ding{72}} & 51.77\textsuperscript{\ding{73}} & 37.95\textsuperscript{\ding{73}} & 21.71\textsuperscript{\ding{73}} \\
    
    SegGPT & 95.77\textsuperscript{\ding{72}} & 81.58\textsuperscript{\ding{72}} & 99.16\textsuperscript{\ding{72}} & 91.16\textsuperscript{\ding{72}} & 65.93\textsuperscript{\ding{73}} & 85.14\textsuperscript{\ding{72}} & 65.22\textsuperscript{\ding{117}} & 50.75\textsuperscript{\ding{117}} & 42.45 & 77.90\textsuperscript{\ding{72}} & 47.10\textsuperscript{\ding{72}} & 38.61\textsuperscript{\ding{72}} \\
    
    \midrule
    \textit{personalized} \\
    PDM & 29.99 & 10.97 & 2.79 & 21.03 & 8.95 & 0.11 & 26.39 & 10.98 & 1.12 & 23.62 & 9.10 & 1.27 \\
    
    PDM+PerSAM  & 50.09 & 60.08 & 33.37 & 64.36 & 53.82 & 41.85 & 35.56 & 45.34 & 22.33 & 28.93 & 25.25 & 11.72 \\
    
    PRPG & - & - & - & 81.52\textsuperscript{\ding{117}} & 37.34 & 68.74\textsuperscript{\ding{73}} & 60.68 & 34.56 & 40.41\textsuperscript{\ding{117}} & - & - & - \\
    
    \midrule
    \textit{generalist} \\
    VP & 24.83 & 18.11 & 0.03 & 38.50 & 14.34 & 4.86 & 17.48 & 12.10 & 0.14 & 8.87 & 4.16 & 0.10 \\
    
    Painter & 56.56 & 51.58 & 29.76 & 72.07 & 49.75 & 56.88\textsuperscript{\ding{117}} & 26.93 & 25.44 & 6.87 & 19.53 & 15.59 & 4.20 \\

    LVM & 43.86 & 33.92 & 19.49 & 54.65 & 27.96 & 30.23 & 22.64 & 13.50 & 2.00 & 16.38 & 8.73 & 1.14 \\

    OmniGen & 33.24 & 37.33 & 9.52 & 44.87  & 41.48 & 18.54 & 20.75 & 20.57 & 2.19 & 13.43 & 14.88 & 1.77 \\
    
    \rowcolor{gray!20}
    PICO (ours) & 90.97\textsuperscript{\ding{73}} & 76.13\textsuperscript{\ding{73}} & 62.82\textsuperscript{\ding{117}} & 71.02 & 54.71\textsuperscript{\ding{117}} & 49.84 & 68.72\textsuperscript{\ding{72}} & 60.26\textsuperscript{\ding{72}} & 44.88\textsuperscript{\ding{73}} & 49.52\textsuperscript{\ding{117}} & 33.63\textsuperscript{\ding{117}} & 14.90\textsuperscript{\ding{117}} \\
    \midrule
     PICO+TTS & 92.04 & 76.85 & 98.14 & 72.33 & 56.54 & 59.62 & 69.90 & 63.60 & 48.50 & 50.66 & 35.23 & 15.96 \\
    $\Delta$ w/ TTS & \deltaup{1.07} & \deltaup{0.12} & \deltaup{35.32} & \deltaup{1.31} & \deltaup{1.83} & \deltaup{9.87} & \deltaup{1.18} & \deltaup{3.34} & \deltaup{3.62} & \deltaup{1.14} & \deltaup{1.60} & \deltaup{1.06} \\
    \bottomrule
  \end{tabular}
  \label{tab:personalized-seg}
  \end{adjustbox}
\end{table}

\begin{table}[t]
\vspace{-3mm}
\centering
\caption{\textbf{Comparison of baseline methods.} (Top) personalized segmentation methods; (Bottom) large-scale pretrained segmentation models.
PICO uses minimal supervision with a diffusion backbone and remains flexible for novel test-time tasks.}
\label{tab:segmentation_comparison}
\vspace{-3mm}
\begin{adjustbox}{width=\linewidth}
\begin{tabular}{@{}lllll@{}}
\toprule
Method & Use of Generative Prior & Features & Seg. Method & Test-time New Instance? \\
\midrule
PDM & Feature extractor & SDXL-turbo~\citep{sauer2024adversarial} & Attention map & \Checkmark \\
PRPG & Synthetic data generator & Personalized DINOv2~\citep{oquab2024dinov2} & Attention map & \XSolidBrush (retraining required) \\
\rowcolor{gray!20}
PICO (ours) & In-context learner & -- & Direct output & \Checkmark \\
\bottomrule
\end{tabular}
\end{adjustbox}

\vspace{0.6em}

\begin{adjustbox}{width=\linewidth}
\begin{tabular}{@{}lllll@{}}
\toprule
Method & Seg. Data / Total Data & Training & Loss & \\ 
\midrule
PerSAM & 11M / 11M  & Finetuned from MAE-pretrained ViT-H~\citep{he2022masked} & Cross-entropy & \\
SegGPT & 254K / 254K & Finetuned from Painter~\citep{wang2023images} & Smooth L1 & \\
Painter & 138K / 192K & Finetuned from MAE-pretrained ViT-Large~\citep{he2022masked} & Smooth L1 & \\
\rowcolor{gray!20}
PICO (ours) & \textbf{40 / 315} & Finetuned from FLUX (DiT-based)~\citep{flux2024} & Flow-matching & \\
\bottomrule
\end{tabular}
\end{adjustbox}
\vspace{-6mm}
\end{table}

\subsection{Personalized Image Segmentation}
\label{main:perseg}

\noindent{\bf{Datasets.}} We evaluate across four personalized segmentation benchmarks: PerSeg~\citep{zhangpersonalize}, DOGS~\citep{sundaram2024personalized}, PODS~\citep{sundaram2024personalized}, and PerMIS~\citep{samuel2024s}. 
While PerSeg and DOGS mainly contain either single instances or distinct instances easily segmented using semantic cues, PODS is more challenging due to variations in viewpoints, scales, and distractors.
PerMIS, sourced from video frames, further increases the difficulty by emphasizing instance-level segmentation.

\noindent{\bf{Baselines.}} 
We compare PICO with three groups of state-of-the-art methods:
(i) Large-scale pretrained segmentors: PerSAM~\citep{zhangpersonalize} and SegGPT~\citep{SegGPT}, both trained on extensive collections of annotation segmentation masks. %
(ii) Personalized representation learners: PDM (diffusion features)~\citep{samuel2024s} and PRPG (personalized features via synthetic-data finetuning)~\citep{sundaram2024personalized}, followed by using attention maps for instance localization. 
(iii) Generalist ICL models: Visual Prompting (VP)~\citep{bar2022visual}, Painter~\citep{wang2023images}, LVM~\citep{bai2024sequential} and OmniGen~\citep{xiao2025omnigen}.  
%

\noindent{\bf{Evaluation Metrics.}}
Following~\citep{samuel2024s, sundaram2024personalized}, we report mIOU, bIOU and F1@$0.50$ scores  over all benchmarks.
All the baseline methods we use its official code base and default settings.

\noindent{\bf{Results.}}
Table~\ref{tab:personalized-seg} shows that PICO outperforms generalist ICL models (VP, Painter, LVM, OmniGen) and personalized representation methods (PDM, PRPG), particularly on the more challenging PODS and PerMIS datasets.
While PRPG achieves competitive results on DOGS, its reliance on per-instance synthetic data generation makes it computationally costly and difficult to scale (see Table~\ref{tab:segmentation_comparison}(Top)). Thus, we omit its results on PerSeg and PerMIS, where over $500$ unique instances are each accompanied by a single reference image.
In contrast, PICO's generative in-context learning paradigm enables instant adaptation to new instances at inference without retraining, offering strong practical advantages.
Notably, compared to large-scale pretrained segmentors, PICO achieves comparable performance while using up to four orders of magnitude fewer labeled data (see Table~\ref{tab:segmentation_comparison}(Bottom)), highlighting its superior data efficiency enabled by generative priors. 
%
%
Qualitative results are provided in Figure~\ref{fig:app:more-results} of the Appendix.

\begin{figure}[!t]
  \centering
  \includegraphics[width=\linewidth]{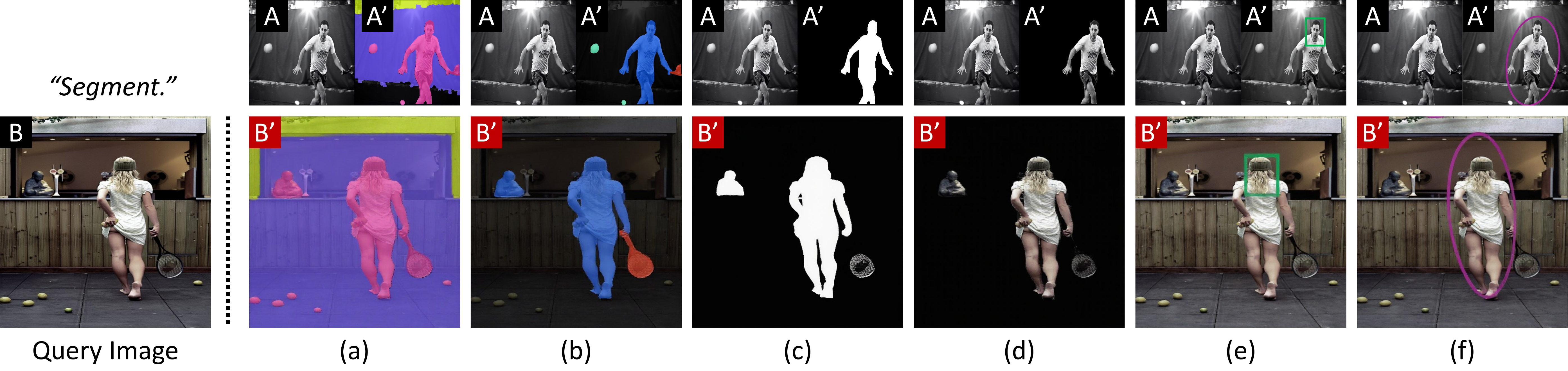}
  \vspace{-6mm}
  \caption{\textbf{Personalized segmentation with visual prompt control.}
    Given the same query image $B$ and text prompt (``Segment''),  PICO produces diverse outputs on $B$ by varying the visual exemplar $(A\!\to\!A')$, controlling task type, style, granularity, and spatial focus.}
  \label{fig:freeform-seg}
  \vspace{-5mm}
\end{figure}

\noindent{\bf{Free-Form Inputs and Task Flexibility.}}
\label{main:freeform-seg}
Beyond dense masks, PICO supports free-form inputs such as sparse annotations (\textit{e.g.}, bounding boxes, circles) or part-level references, offering greater flexibility for personalization. 
As shown in Figure~\ref{fig:freeform-seg}, PICO generates diverse segmentation outputs conditioned on visual exemplars while keeping text prompts fixed. 
Outputs vary along multiple dimensions: 
(i) \textit{Task type}: stuff (a) vs.\ semantic (b) segmentation with arbitrary color coding and transparency; 
(ii) \textit{Style}: binary silhouettes (c) vs.\ matting-like masks (d); 
(iii) \textit{Granularity}: dense masks vs.\ sparse annotations; 
(iv) \textit{Spatial focus}: whole-object (f) vs.\ part-level regions (e). 
PICO reliably aligns with the intent, style, and semantics conveyed in visual prompts, which are often hard to specify in text. 
Additional analyses of visual prompt effects are in Appendix~\ref{abl_img}.

%

\begin{figure}[!b]
  \centering
  \vspace{-5mm}
  \includegraphics[width=\linewidth]{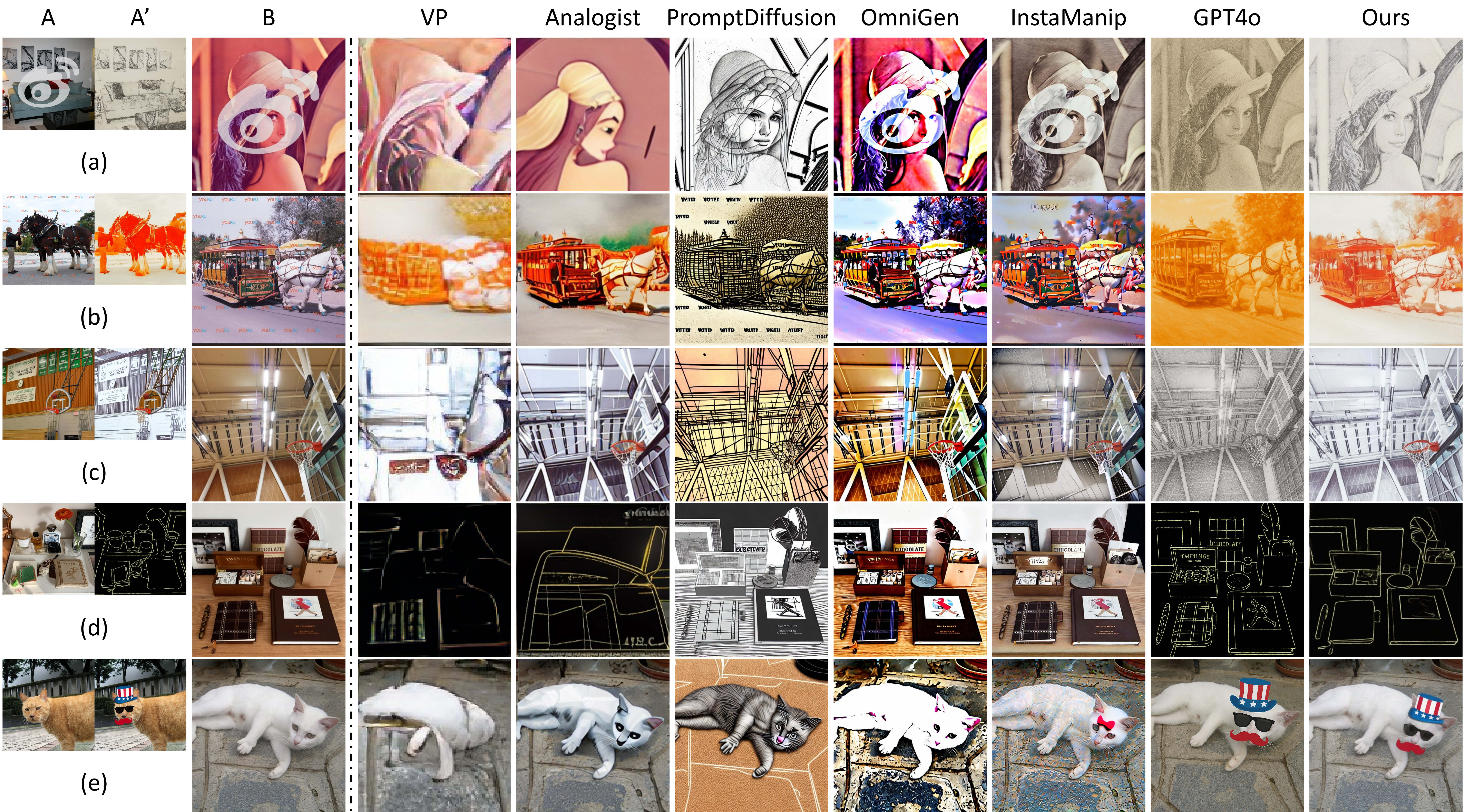}
  \caption{\textbf{Qualitative comparisons on test-time personalized tasks.} Each task is defined by a visual exemplar $(A \!\to\! A')$. We compare PICO with five representative baselines on: (a)(b) watermark removal + style transfer; (c) background-only stylization; (d) contour-only edge detection; and (e) sticker insertion.}
  \label{fig:novel}
\end{figure}

\subsection{Personalized Test-time Task Generalization}
\label{main:test-time}

\noindent{\bf{Task Definition.}}
We evaluate test-time personalization on user-defined visual tasks that differ from conventional CV setups. Specifically, we focus on:
(i) \textbf{Composite tasks} requiring multi-step operations (\textit{e.g.}, watermark removal followed by stylization).
(ii) \textbf{Spatially constrained tasks}, traditionally performed globally but here applied locally or selectively (\textit{e.g.}, contour-only edge detection, background-only stylization).
(iii) \textbf{Semantic-conditional tasks} demanding context-aware edits (\textit{e.g.}, adding stickers to semantically relevant image regions).

\noindent{\bf{Baselines.}}
Given these novel tasks, we compare PICO with representative state-of-the-art methods supporting visual instructions, including: 
(i) Inference-based method: VP~\citep{bar2022visual}, Analogist~\citep{gu2024analogist};
(ii) Training-based method: PromptDiffusion~\citep{wang2023context}, LVM~\citep{bai2024sequential}, OmniGen~\citep{xiao2025omnigen}, InstaManip~\citep{lai2024unleashing};
(iii) \textit{Commercial multimodal models}: GPT-4o~\citep{gpt4o2024}. 
%
Textual instructions for these methods follow Analogist’s GPT-4o-based reasoning procedure.

%
\noindent{\bf{Evaluation Metrics}}
We evaluate two representative composite tasks with clear quantitative protocols:  
(1) Deraining with inpainting, measured by PSNR and SSIM against clean ground truth.  
(2) Inpainting with stylization, measured by Gram distance for style fidelity, LPIPS for content preservation, and ArtFID~\citep{chung2024style} for overall perceptual quality.  
Full protocols and dataset details are provided in the Appendix~\ref{app:quantitative-details}.

\begin{table}[t]
\caption{\textbf{Quantitative comparison on test-time personalized tasks.}
The best results are in \textbf{bold}, second-best are \underline{underlined}. GPT-4o* results are based on 10 random samples due to API constraints.}
\vspace{-1mm}
\centering
\begin{adjustbox}{width=\linewidth}
\begin{tabular}{@{}l|c|ccccccc>{\columncolor[gray]{0.9}}c@{}}
\toprule
  & Ref & OmniGen & LVM & VP & Analogist & PromptDiff & InstaManip & GPT-4o* & PICO (Ours) \\
\midrule
\multicolumn{10}{@{}l}{\textit{deraining with inpainting}} \\
PSNR (dB)$\uparrow$ & $\infty$ & \underline{15.63} & 15.39 & 14.62 & 12.35 & 9.64 & 10.94 & 12.29 & \textbf{22.24} \\
SSIM $\uparrow$    & $1.0$    & \underline{0.47} & 0.35  & 0.36  & 0.35  & 0.10 & 0.33  & 0.26  & \textbf{0.67} \\
\midrule
\multicolumn{10}{@{}l}{\textit{inpainting with stylization}} \\
Gram$\downarrow$   & 17.29 & 90.78 & 27.11 & 28.96 & 26.53 & 61.61 & 44.39 & \underline{22.04} & \textbf{21.27} \\
FID$\downarrow$    & 1.71  & 1.92  & 1.90  & \underline{1.86} & \textbf{1.82} & \underline{1.86} & 1.88 & 1.87 & 1.87 \\
LPIPS$\downarrow$  & 0.62  & \underline{0.59} & 0.61 & 0.82 & 0.70 & 0.77 & \underline{0.60} & 0.68 & \textbf{0.52} \\
ArtFID$\downarrow$ & 4.38  & 4.63  & 4.68  & 5.19 & 4.79 & 5.06 & \underline{4.59} & 4.81 & \textbf{4.38} \\
\bottomrule
\end{tabular}
\end{adjustbox}
\label{tab:novel_std_metric}
\vspace{-1mm}
\end{table}

\begin{figure}[t]
\vspace{-1mm}
  \centering
  \begin{minipage}[t]{0.49\textwidth}
    \vspace{0pt}%
    \captionsetup{type=table,width=\linewidth,justification=centering}
    \captionof{table}{\textbf{Quantitative results on w/wo texts.}}
    \vspace{-0.9em}
    \label{tbl:ablation_text}
    \scriptsize
    \setlength{\tabcolsep}{5.5pt}
    \renewcommand{\arraystretch}{1.1}

    \begin{tabular*}{\linewidth}{@{\extracolsep{\fill}} l|ccc @{}}
      \toprule
      Method           & Pers.\ Seg$\uparrow$ & Normal$\downarrow$ & Z-depth$\downarrow$ \\
      \midrule
      VTM (10-Shot)    & --     & 11.4391 & \textbf{0.0316} \\
      \midrule
      Ours w/o Text    & 66.88  & 12.7105 & 0.0432 \\
      \rowcolor{gray!15}
      Ours w Text      & \textbf{68.72} & \textbf{10.5306} & 0.0377 \\
      \bottomrule
    \end{tabular*}

    \vspace{0.9em} 

    \begin{tabular*}{\linewidth}{@{\extracolsep{\fill}} l|ccc @{}}
      \toprule
      Method           & 2DEdge$\downarrow$ & 2DKeypoint$\downarrow$ & Reshading$\downarrow$ \\
      \midrule
      VTM (10-Shot)    & 0.0791 & 0.0639 & \textbf{0.1089} \\
      \midrule
      Ours w/o Text    & 0.0538 & 0.0609 & 0.1518 \\
      \rowcolor{gray!15}
      Ours w Text      & \textbf{0.0515} & \textbf{0.0497} & 0.1364 \\
      \bottomrule
    \end{tabular*}
  \end{minipage}%
  \hfill
  \begin{minipage}[t]{0.49\textwidth}
    \vspace{0pt}%
    \centering
    \includegraphics[width=\linewidth,keepaspectratio]{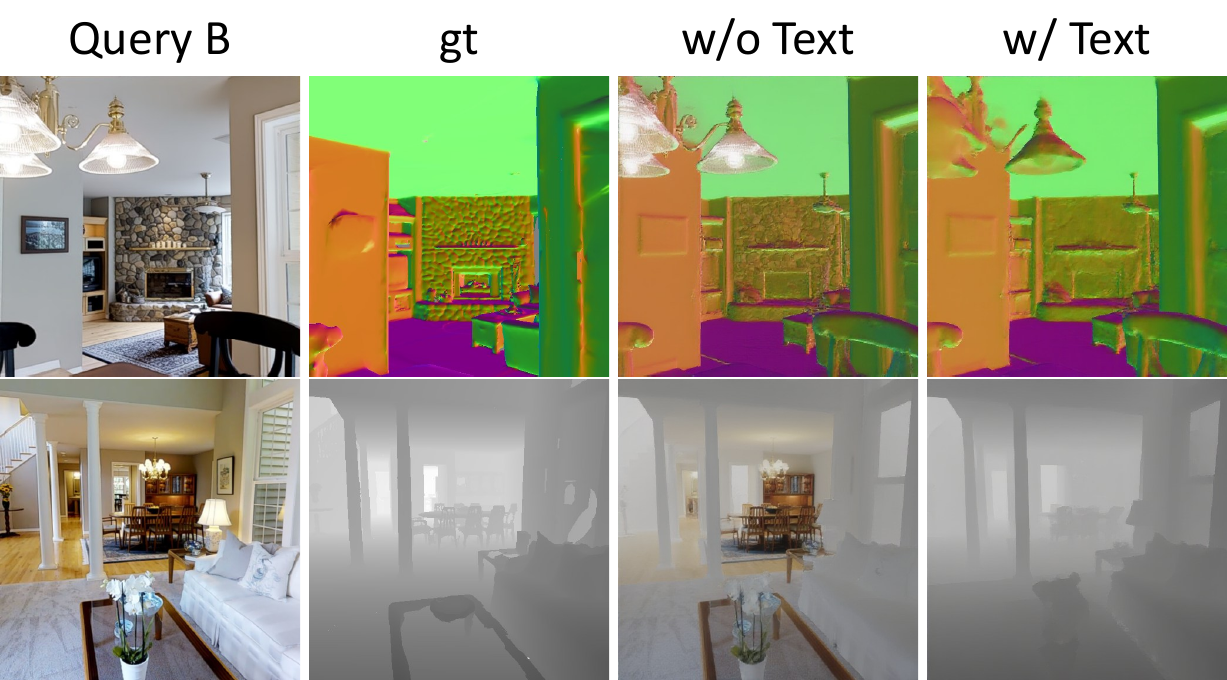}
    \captionsetup{type=figure}
    \vspace{-1.8em}
    \captionof{figure}{\textbf{Qualitative results on w/wo texts.}}
    \label{fig:ablation_text}
  \end{minipage}
\vspace{-5mm}
\end{figure}

\noindent{\bf{Results.}}
Figure~\ref{fig:novel} shows that PICO effectively handles diverse test-time defined novel tasks, clearly surpassing all baselines. 
%
%
Training-based methods (PromptDiffusion, OmniGen, InstaManip)  primarily target semantic-driven editing and struggle to match exemplar  appearances, especially for non-RGB outputs (\textit{e.g.}, edge maps in Figure~\ref{fig:novel}(d)).
%
%
Inference-based methods (VP, Analogist) can mimic target transformations roughly, but suffer from low fidelity and noticeable visual artifacts.
GPT-4o~\citep{gpt4o2024} shows promising in-context reasoning, but two  major limitations are observed:
(1) Spatial misalignment: While semantic content is preserved, pixel layouts are distorted, harming precision tasks (see Figure~\ref{fig:novel}(d-e)).
(2) Over-reliance on abstract semantics:
outputs rely on abstract semantics rather than exemplar fidelity, producing generic effect ( ``sketch'' or ``orange-tone'') in stylization tasks (see Figure~\ref{fig:novel}(a-c)).
In contrast, PICO produces outputs consistently aligned with exemplar cues in both spatial detail and semantic fidelity, demonstrating robust visual reasoning.
Table~\ref{tab:novel_std_metric} confirms this quantitatively, with PICO consistently achieving the best results across both tasks and metrics.
Additional qualitative comparisons are provided in Appendix~\ref{app:qualitative-comp}.


\subsection{Ablation Studies}

\noindent{\bf{Effects of Text Prompts.}}
We first quantify the importance of minimal textual prompts in disambiguating multiple visual tasks. Specifically, we evaluate on personalized segmentation (PODS) and five dense prediction tasks from Taskonomy~\citep{zamir2018taskonomy}, using $1{,}000$ test samples per task.
%
Metrics follow~\citep{kim2023universal}: mean error (mErr) for surface normal, and RMSE for others.Predictions are converted from RGB to raw output space before scoring.

Table~\ref{tbl:ablation_text} shows that adding text prompts consistently improves performance, acting as soft task boundaries that reduce ambiguity beyond visual prompts alone. 
Figure~\ref{fig:ablation_text} illustrates typical confusions without text, such as RGB-like outputs instead of surface normal maps. 
With text, the model cleanly separates task outputs.  
For reference, we include VTM~\citep{kim2023universal}, a state-of-the-art 10-shot fine-tuning method for dense prediction. 
Remarkably, our generative in-context learner surpasses this specialized approach on tasks such as surface normal estimation and texture edge detection, highlighting strong generalization and data efficiency enabled by generative priors. 




%
\begin{figure}[t]
  \centering

  \begin{minipage}[t]{0.60\linewidth}\vspace{0pt}%
    \centering
    \includegraphics[height=4.6cm]{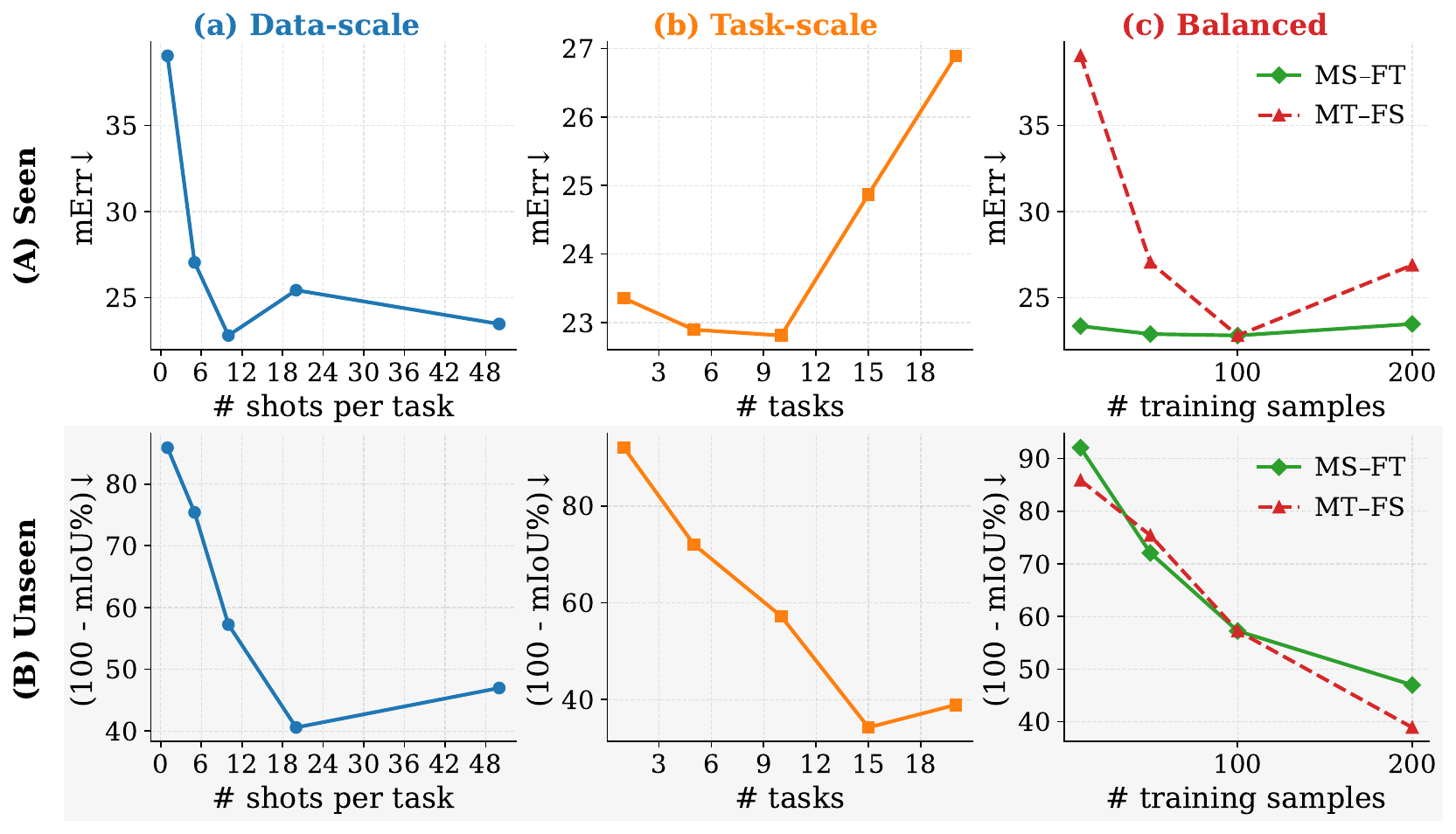}
    \captionsetup{type=figure,width=\linewidth,justification=centering}
    \captionof{figure}{\textbf{Quantitative comparisons of different scaling strategies.} Lower values indicate better performance.}
    \label{fig:abl_sn}
  \end{minipage}
  \hfill
  \begin{minipage}[t]{0.38\linewidth}\vspace{0pt}%
    \centering
    \includegraphics[height=4.6cm]{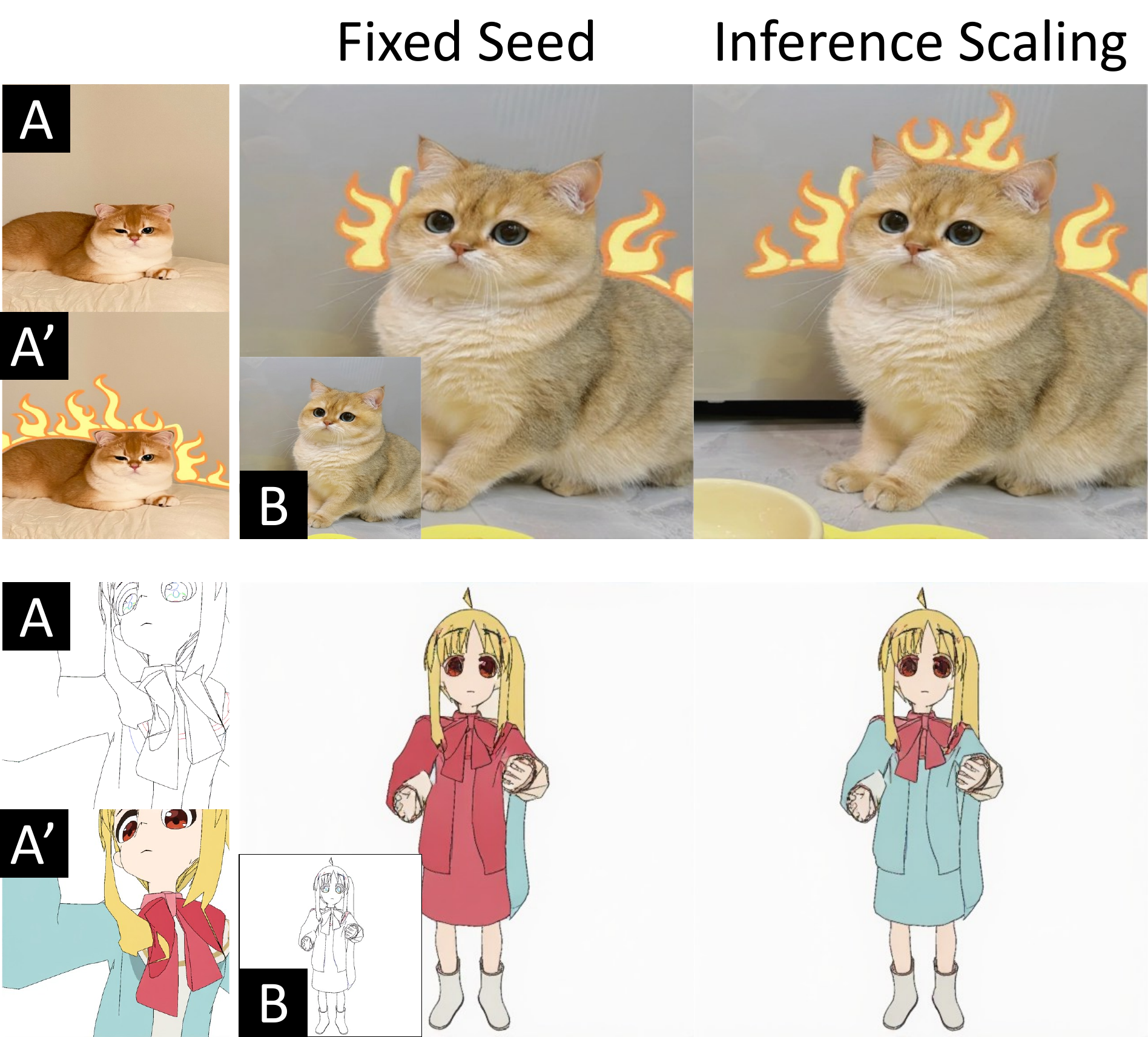}
    \captionsetup{type=figure,width=\linewidth,justification=centering}
    \captionof{figure}{\textbf{Test-time scaling.} Results w/wo our attn-guided seed scorer.}
    \label{fig:abl_tts}
  \end{minipage}
\vspace{-5mm}
\end{figure}

\noindent{\bf{Task vs. Data Scaling.}}
We systematically study how task diversity and data volume shape model generalization. With LoRA rank fixed ($r{=}128$) and $10k$ training steps, we evaluate three settings:
%
(i) \textbf{Data-scale sweep: }fix 10 tasks, vary shots per task: ($K \in {1, 5, 10, 20, 50}$).
(ii) \textbf{Task-scale sweep: }fix 10 shots, vary number of tasks ($N \in {1, 5, 10, 15, 20}$).
(iii) \textbf{Balanced sweep:} fix total training images (${10, 50, 100, 200}$), compare many-tasks–few-shots ($N > K$) against few-tasks–many-shots ($N < K$) regimes.
%
%
We evaluate on both in-domain tasks seen during training (\textit{e.g.}, surface normal estimation) and out-of-domain tasks not seen during training (\textit{e.g.}, personalized segmentation).

Results are shown in Figure~\ref{fig:abl_sn}. 
For in-domain tasks, more data volume consistently improves performance (Figure.\ref{fig:abl_sn}A-a), while adding tasks hurts (Figure.\ref{fig:abl_sn}A-b), indicating limited capacity for memorizing multiple tasks.
%
%
For out-of-domain generalization, performance improves with more data per task only up to 20 shots, after which it declines due to over-specialization (Figure.\ref{fig:abl_sn}B-a).
In contrast, task diversity consistently boosts generalization (Figure.\ref{fig:abl_sn}B-b). 
Under fixed budgets, the many-tasks–few-shots strategy increasingly outperforms fewer-tasks–many-shots as task count grows (Figure.\ref{fig:abl_sn}B-c).
These results support our \emph{visual-relation–space} hypothesis: \textbf{data scaling helps memorization of seen tasks, whereas task diversity is key for robust generalization to novel, user-defined tasks.}

\noindent{\bf{Test-Time Scaling.}}
We evaluate our early-step seed-scoring strategy on personalized segmentation tasks and observe consistent improvements across all dataset (see Table~\ref{tab:personalized-seg}).
Qualitative examples in Figure~\ref{fig:abl_tts} further show that scaled outputs align more faithfully with visual exemplars. 
Additional ablations are provided in Appendix~\ref{app:ablations}.

\section{Conclusion}
\label{dis}

In this paper, we present PICO, a novel approach for personalized vision by reformulating it as a visual in-context learning (ICL) problem. 
Unlike existing methods that rely heavily on task-specific fine-tuning or synthetic data augmentation, PICO leverages a unified visual-relation space, enabling pretrained diffusion models to interpret user-defined tasks from a single visual demonstration at inference. 
Extensive experiments show that PICO adapts flexibly to novel objects and tasks, achieving strong performance across recognition and generation, and highlighting the potential of generative models as versatile visual in-context reasoners.

\noindent{\bf{Limitation and Future Work.}}
\label{limitation}
PICO generalizes well within the trained visual-relation space but is less reliable on entirely novel task types outside it. 
This aligns with human learning, \textit{i.e.}, people extrapolate best within familiar domains, but broadening the method to truly novel tasks remains an open challenge.
The four-panel input format, while effective, inherently limits the number and richness of demonstrations. 
Future work includes extending PICO to richer or sequential context (\textit{e.g.}, videos or long-context models~\citep{gu2025long}) to broaden task coverage and strengthen visual reasoning.


\newpage
\bibliography{iclr2026_conference}

\begin{thebibliography}{62}
\providecommand{\natexlab}[1]{#1}
\providecommand{\url}[1]{\texttt{#1}}
\expandafter\ifx\csname urlstyle\endcsname\relax
  \providecommand{\doi}[1]{doi: #1}\else
  \providecommand{\doi}{doi: \begingroup \urlstyle{rm}\Url}\fi

\bibitem[Agustsson \& Timofte(2017)Agustsson and Timofte]{Agustsson_2017_CVPR_Workshops}
Eirikur Agustsson and Radu Timofte.
\newblock {NTIRE} 2017 challenge on single image super-resolution: Dataset and study.
\newblock In \emph{CVPRW}, 2017.

\bibitem[Alaluf et~al.(2024)Alaluf, Richardson, Tulyakov, Aberman, and Cohen-Or]{alaluf2024myvlm}
Yuval Alaluf, Elad Richardson, Sergey Tulyakov, Kfir Aberman, and Daniel Cohen-Or.
\newblock {MyVLM}: Personalizing vlms for user-specific queries.
\newblock In \emph{ECCV}, 2024.

\bibitem[Ancuti et~al.(2019)Ancuti, Ancuti, Sbert, and Timofte]{Dense-Haze_2019}
Codruta~O. Ancuti, Cosmin Ancuti, Mateu Sbert, and Radu Timofte.
\newblock Dense haze: A benchmark for image dehazing with dense-haze and haze-free images.
\newblock In \emph{ICIP}, 2019.

\bibitem[Bai et~al.(2024)Bai, Geng, Mangalam, Bar, Yuille, Darrell, Malik, and Efros]{bai2024sequential}
Yutong Bai, Xinyang Geng, Karttikeya Mangalam, Amir Bar, Alan~L Yuille, Trevor Darrell, Jitendra Malik, and Alexei~A Efros.
\newblock Sequential modeling enables scalable learning for large vision models.
\newblock In \emph{CVPR}, 2024.

\bibitem[Bar et~al.(2022)Bar, Gandelsman, Darrell, Globerson, and Efros]{bar2022visual}
Amir Bar, Yossi Gandelsman, Trevor Darrell, Amir Globerson, and Alexei Efros.
\newblock Visual prompting via image inpainting.
\newblock \emph{NeurIPS}, 2022.

\bibitem[Brown et~al.(2020)Brown, Mann, Ryder, Subbiah, Kaplan, Dhariwal, Neelakantan, Shyam, Sastry, Askell, et~al.]{brown2020language}
Tom Brown, Benjamin Mann, Nick Ryder, Melanie Subbiah, Jared~D Kaplan, Prafulla Dhariwal, Arvind Neelakantan, Pranav Shyam, Girish Sastry, Amanda Askell, et~al.
\newblock Language models are few-shot learners.
\newblock \emph{NeurIPS}, 2020.

\bibitem[Chung et~al.(2024)Chung, Hyun, and Heo]{chung2024style}
Jiwoo Chung, Sangeek Hyun, and Jae-Pil Heo.
\newblock Style injection in diffusion: A training-free approach for adapting large-scale diffusion models for style transfer.
\newblock In \emph{CVPR}, 2024.

\bibitem[Cohen et~al.(2022)Cohen, Gal, Meirom, Chechik, and Atzmon]{cohen2022my}
Niv Cohen, Rinon Gal, Eli~A Meirom, Gal Chechik, and Yuval Atzmon.
\newblock “{This is my unicorn, Fluffy}”: Personalizing frozen vision-language representations.
\newblock In \emph{ECCV}, 2022.

\bibitem[Crawford(2019)]{crawfordcats}
Nick Crawford.
\newblock Cat dataset.
\newblock \url{https://www.kaggle.com/datasets/crawford/cat-dataset}, 2019.

\bibitem[Dai et~al.(2024)Dai, Zhou, Li, Li, and Loy]{InclusionMatching2024}
Yuekun Dai, Shangchen Zhou, Qinyue Li, Chongyi Li, and Chen~Change Loy.
\newblock Learning inclusion matching for animation paint bucket colorization.
\newblock \emph{CVPR}, 2024.

\bibitem[Dong et~al.(2024)Dong, Li, Dai, Zheng, Ma, Li, Xia, Xu, Wu, Chang, et~al.]{dong2024survey}
Qingxiu Dong, Lei Li, Damai Dai, Ce~Zheng, Jingyuan Ma, Rui Li, Heming Xia, Jingjing Xu, Zhiyong Wu, Baobao Chang, et~al.
\newblock A survey on in-context learning.
\newblock In \emph{Proceedings of the 2024 Conference on Empirical Methods in Natural Language Processing}, 2024.

\bibitem[Dosovitskiy et~al.(2021)Dosovitskiy, Beyer, Kolesnikov, Weissenborn, Zhai, Unterthiner, Dehghani, Minderer, Heigold, Gelly, Uszkoreit, and Houlsby]{dosovitskiy2020vit}
Alexey Dosovitskiy, Lucas Beyer, Alexander Kolesnikov, Dirk Weissenborn, Xiaohua Zhai, Thomas Unterthiner, Mostafa Dehghani, Matthias Minderer, Georg Heigold, Sylvain Gelly, Jakob Uszkoreit, and Neil Houlsby.
\newblock {An Image is Worth 16x16 Words:} transformers for image recognition at scale.
\newblock \emph{ICLR}, 2021.

\bibitem[Esser et~al.(2024)Esser, Kulal, Blattmann, Entezari, M{\"u}ller, Saini, Levi, Lorenz, Sauer, Boesel, et~al.]{esser2024scaling}
Patrick Esser, Sumith Kulal, Andreas Blattmann, Rahim Entezari, Jonas M{\"u}ller, Harry Saini, Yam Levi, Dominik Lorenz, Axel Sauer, Frederic Boesel, et~al.
\newblock Scaling rectified flow transformers for high-resolution image synthesis.
\newblock In \emph{ICML}, 2024.

\bibitem[Fang et~al.(2018)Fang, Wu, Yang, Savarese, and Lim]{demo2vec2018cvpr}
Kuan Fang, Te-Lin Wu, Daniel Yang, Silvio Savarese, and Joseph~J. Lim.
\newblock Demo2vec: Reasoning object affordances from online videos.
\newblock In \emph{CVPR}, 2018.

\bibitem[Fu et~al.(2024)Fu, Yin, Hu, Wang, Ma, Tan, Shen, Lin, and Long]{fu2024geowizard}
Xiao Fu, Wei Yin, Mu~Hu, Kaixuan Wang, Yuexin Ma, Ping Tan, Shaojie Shen, Dahua Lin, and Xiaoxiao Long.
\newblock {GeoWizard}: Unleashing the diffusion priors for 3d geometry estimation from a single image.
\newblock In \emph{ECCV}, 2024.

\bibitem[Gal et~al.(2023)Gal, Alaluf, Atzmon, Patashnik, Bermano, Chechik, and Cohen-or]{galimage}
Rinon Gal, Yuval Alaluf, Yuval Atzmon, Or~Patashnik, Amit~Haim Bermano, Gal Chechik, and Daniel Cohen-or.
\newblock An image is worth one word: Personalizing text-to-image generation using textual inversion.
\newblock In \emph{ICLR}, 2023.

\bibitem[Gu et~al.(2025)Gu, Mao, and Shou]{gu2025long}
Yuchao Gu, Weijia Mao, and Mike~Zheng Shou.
\newblock Long-context autoregressive video modeling with next-frame prediction.
\newblock \emph{arXiv preprint arXiv:2503.19325}, 2025.

\bibitem[Gu et~al.(2024)Gu, Yang, Liao, Huo, and Gao]{gu2024analogist}
Zheng Gu, Shiyuan Yang, Jing Liao, Jing Huo, and Yang Gao.
\newblock Analogist: Out-of-the-box visual in-context learning with image diffusion model.
\newblock \emph{TOG}, 2024.

\bibitem[He et~al.(2025{\natexlab{a}})He, Shen, Fang, Xiao, Tang, Zhang, Zuo, Guo, and Li]{he2025diffusion}
Chunming He, Yuqi Shen, Chengyu Fang, Fengyang Xiao, Longxiang Tang, Yulun Zhang, Wangmeng Zuo, Zhenhua Guo, and Xiu Li.
\newblock Diffusion models in low-level vision: A survey.
\newblock \emph{IEEE TPAMI}, 2025{\natexlab{a}}.

\bibitem[He et~al.(2025{\natexlab{b}})He, Li, Yin, Liang, Li, Zhou, Zhang, Liu, and Chen]{he2024lotus}
Jing He, Haodong Li, Wei Yin, Yixun Liang, Leheng Li, Kaiqiang Zhou, Hongbo Zhang, Bingbing Liu, and Ying-Cong Chen.
\newblock Lotus: Diffusion-based visual foundation model for high-quality dense prediction.
\newblock In \emph{ICLR}, 2025{\natexlab{b}}.

\bibitem[He et~al.(2022)He, Chen, Xie, Li, Doll{\'a}r, and Girshick]{he2022masked}
Kaiming He, Xinlei Chen, Saining Xie, Yanghao Li, Piotr Doll{\'a}r, and Ross Girshick.
\newblock Masked autoencoders are scalable vision learners.
\newblock In \emph{CVPR}, 2022.

\bibitem[Hu et~al.(2022)Hu, Shen, Wallis, Allen-Zhu, Li, Wang, Wang, Chen, et~al.]{hu2022lora}
Edward~J Hu, Yelong Shen, Phillip Wallis, Zeyuan Allen-Zhu, Yuanzhi Li, Shean Wang, Lu~Wang, Weizhu Chen, et~al.
\newblock {LoRA}: Low-rank adaptation of large language models.
\newblock \emph{ICLR}, 2022.

\bibitem[Huang et~al.(2024)Huang, Wang, Wu, Shi, Dou, Liang, Feng, Liu, and Zhou]{huang2024context}
Lianghua Huang, Wei Wang, Zhi-Fan Wu, Yupeng Shi, Huanzhang Dou, Chen Liang, Yutong Feng, Yu~Liu, and Jingren Zhou.
\newblock In-context lora for diffusion transformers.
\newblock \emph{arXiv preprint arXiv:2410.23775}, 2024.

\bibitem[Huang et~al.(2025)Huang, Song, Zhang, Guo, Wang, Shou, and Liu]{huang2025photodoodle}
Shijie Huang, Yiren Song, Yuxuan Zhang, Hailong Guo, Xueyin Wang, Mike~Zheng Shou, and Jiaming Liu.
\newblock {PhotoDoodle}: Learning artistic image editing from few-shot pairwise data.
\newblock \emph{arXiv preprint arXiv:2502.14397}, 2025.

\bibitem[Jiang et~al.(2020)Jiang, Wang, Yi, Chen, Huang, Luo, Ma, and Jiang]{Kui_2020_CVPR}
Kui Jiang, Zhongyuan Wang, Peng Yi, Chen Chen, Baojin Huang, Yimin Luo, Jiayi Ma, and Junjun Jiang.
\newblock Multi-scale progressive fusion network for single image deraining.
\newblock In \emph{CVPR}, 2020.

\bibitem[Jiang et~al.(2025)Jiang, Jiang, Yang, Liu, Tsang, and Shou]{jiang2025balanced}
Yuxin Jiang, Liming Jiang, Shuai Yang, Jia-Wei Liu, Ivor Tsang, and Mike~Zheng Shou.
\newblock Balanced image stylization with style matching score.
\newblock In \emph{ICCV}, 2025.

\bibitem[Ke et~al.(2024)Ke, Obukhov, Huang, Metzger, Daudt, and Schindler]{ke2023repurposing}
Bingxin Ke, Anton Obukhov, Shengyu Huang, Nando Metzger, Rodrigo~Caye Daudt, and Konrad Schindler.
\newblock Repurposing diffusion-based image generators for monocular depth estimation.
\newblock In \emph{CVPR}, 2024.

\bibitem[Kim et~al.(2023)Kim, Kim, Cho, Luo, and Hong]{kim2023universal}
Donggyun Kim, Jinwoo Kim, Seongwoong Cho, Chong Luo, and Seunghoon Hong.
\newblock Universal few-shot learning of dense prediction tasks with visual token matching.
\newblock In \emph{ICLR}, 2023.

\bibitem[Kirillov et~al.(2023)Kirillov, Mintun, Ravi, Mao, Rolland, Gustafson, Xiao, Whitehead, Berg, Lo, et~al.]{kirillov2023segment}
Alexander Kirillov, Eric Mintun, Nikhila Ravi, Hanzi Mao, Chloe Rolland, Laura Gustafson, Tete Xiao, Spencer Whitehead, Alexander~C Berg, Wan-Yen Lo, et~al.
\newblock Segment anything.
\newblock In \emph{ICCV}, 2023.

\bibitem[Labs(2024)]{flux2024}
Black~Forest Labs.
\newblock Flux.
\newblock \url{https://github.com/black-forest-labs/flux}, 2024.

\bibitem[Lai et~al.(2025)Lai, Juefei-Xu, Liu, Dai, Mehta, Zhu, Huang, Rehg, Lee, Zhang, et~al.]{lai2024unleashing}
Bolin Lai, Felix Juefei-Xu, Miao Liu, Xiaoliang Dai, Nikhil Mehta, Chenguang Zhu, Zeyi Huang, James~M Rehg, Sangmin Lee, Ning Zhang, et~al.
\newblock Unleashing in-context learning of autoregressive models for few-shot image manipulation.
\newblock In \emph{CVPR}, 2025.

\bibitem[Lin et~al.(2014)Lin, Maire, Belongie, Hays, Perona, Ramanan, Doll{\'a}r, and Zitnick]{lin2014microsoft}
Tsung-Yi Lin, Michael Maire, Serge Belongie, James Hays, Pietro Perona, Deva Ramanan, Piotr Doll{\'a}r, and C~Lawrence Zitnick.
\newblock Microsoft {COCO}: Common objects in context.
\newblock In \emph{ECCV}, 2014.

\bibitem[Lomonaco \& Maltoni(2017)Lomonaco and Maltoni]{lomonaco2017core50}
Vincenzo Lomonaco and Davide Maltoni.
\newblock {CORe50}: a new dataset and benchmark for continuous object recognition.
\newblock In \emph{CoRL}, 2017.

\bibitem[Lugmayr et~al.(2022)Lugmayr, Danelljan, Romero, Yu, Timofte, and Van~Gool]{lugmayr2022repaint}
Andreas Lugmayr, Martin Danelljan, Andres Romero, Fisher Yu, Radu Timofte, and Luc Van~Gool.
\newblock {RePaint}: Inpainting using denoising diffusion probabilistic models.
\newblock In \emph{CVPR}, 2022.

\bibitem[Mishchenko \& Defazio(2024)Mishchenko and Defazio]{mishchenko2024prodigy}
Konstantin Mishchenko and Aaron Defazio.
\newblock Prodigy: An expeditiously adaptive parameter-free learner.
\newblock In \emph{ICML}, 2024.

\bibitem[Nguyen et~al.(2023)Nguyen, Li, Ojha, and Lee]{nguyen2023visual}
Thao Nguyen, Yuheng Li, Utkarsh Ojha, and Yong~Jae Lee.
\newblock Visual instruction inversion: Image editing via image prompting.
\newblock \emph{NeurIPS}, 2023.

\bibitem[Nguyen et~al.(2024)Nguyen, Liu, Li, Cai, Ojha, and Lee]{nguyenyo}
Thao Nguyen, Haotian Liu, Yuheng Li, Mu~Cai, Utkarsh Ojha, and Yong~Jae Lee.
\newblock {Yo'LLaVA}: Your personalized language and vision assistant.
\newblock \emph{NeurIPS}, 2024.

\bibitem[OpenAI(2024)]{gpt4o2024}
OpenAI.
\newblock Hello {GPT}-4o.
\newblock \url{https://cdn.openai.com/gpt-4o-system-card.pdf}, 2024.

\bibitem[Oquab et~al.(2024)Oquab, Darcet, Moutakanni, Vo, Szafraniec, Khalidov, Fernandez, Haziza, Massa, El-Nouby, et~al.]{oquab2024dinov2}
Maxime Oquab, Timoth{\'e}e Darcet, Th{\'e}o Moutakanni, Huy Vo, Marc Szafraniec, Vasil Khalidov, Pierre Fernandez, Daniel Haziza, Francisco Massa, Alaaeldin El-Nouby, et~al.
\newblock {DINOv2}: Learning robust visual features without supervision.
\newblock \emph{TMLR}, 2024.

\bibitem[Peebles \& Xie(2023)Peebles and Xie]{peebles2023scalable}
William Peebles and Saining Xie.
\newblock Scalable diffusion models with transformers.
\newblock In \emph{ICCV}, 2023.

\bibitem[Qin et~al.(2022)Qin, Dai, Hu, Fan, Shao, and Gool]{qin2022}
Xuebin Qin, Hang Dai, Xiaobin Hu, Deng-Ping Fan, Ling Shao, and Luc~Van Gool.
\newblock Highly accurate dichotomous image segmentation.
\newblock In \emph{ECCV}, 2022.

\bibitem[Radford et~al.(2021)Radford, Kim, Hallacy, Ramesh, Goh, Agarwal, Sastry, Askell, Mishkin, Clark, et~al.]{radford2021learning}
Alec Radford, Jong~Wook Kim, Chris Hallacy, Aditya Ramesh, Gabriel Goh, Sandhini Agarwal, Girish Sastry, Amanda Askell, Pamela Mishkin, Jack Clark, et~al.
\newblock Learning transferable visual models from natural language supervision.
\newblock In \emph{ICML}, 2021.

\bibitem[Rombach et~al.(2022)Rombach, Blattmann, Lorenz, Esser, and Ommer]{rombach2022high}
Robin Rombach, Andreas Blattmann, Dominik Lorenz, Patrick Esser, and Bj{\"o}rn Ommer.
\newblock High-resolution image synthesis with latent diffusion models.
\newblock In \emph{CVPR}, 2022.

\bibitem[Ruiz et~al.(2023)Ruiz, Li, Jampani, Pritch, Rubinstein, and Aberman]{ruiz2023dreambooth}
Nataniel Ruiz, Yuanzhen Li, Varun Jampani, Yael Pritch, Michael Rubinstein, and Kfir Aberman.
\newblock {DreamBooth}: Fine tuning text-to-image diffusion models for subject-driven generation.
\newblock In \emph{CVPR}, 2023.

\bibitem[Samuel et~al.(2024)Samuel, Ben-Ari, Levy, Darshan, and Chechik]{samuel2024s}
Dvir Samuel, Rami Ben-Ari, Matan Levy, Nir Darshan, and Gal Chechik.
\newblock {Where's Waldo:} diffusion features for personalized segmentation and retrieval.
\newblock \emph{NeurIPS}, 2024.

\bibitem[Sauer et~al.(2024)Sauer, Lorenz, Blattmann, and Rombach]{sauer2024adversarial}
Axel Sauer, Dominik Lorenz, Andreas Blattmann, and Robin Rombach.
\newblock Adversarial diffusion distillation.
\newblock In \emph{ECCV}, 2024.

\bibitem[Song et~al.(2025)Song, Liu, and Shou]{song2025omniconsistency}
Yiren Song, Cheng Liu, and Mike~Zheng Shou.
\newblock Omniconsistency: Learning style-agnostic consistency from paired stylization data.
\newblock \emph{NeurIPS}, 2025.

\bibitem[Sundaram et~al.(2025)Sundaram, Chae, Tian, Beery, and Isola]{sundaram2024personalized}
Shobhita Sundaram, Julia Chae, Yonglong Tian, Sara Beery, and Phillip Isola.
\newblock Personalized representation from personalized generation.
\newblock In \emph{ICLR}, 2025.

\bibitem[Toschi et~al.(2023)Toschi, De~Matteo, Spezialetti, De~Gregorio, Di~Stefano, and Salti]{Toschi_2023_CVPR}
Marco Toschi, Riccardo De~Matteo, Riccardo Spezialetti, Daniele De~Gregorio, Luigi Di~Stefano, and Samuele Salti.
\newblock {ReLight My NeRF}: A dataset for novel view synthesis and relighting of real world objects.
\newblock In \emph{CVPR}, 2023.

\bibitem[Wang et~al.(2023{\natexlab{a}})Wang, Wang, Cao, Shen, and Huang]{wang2023images}
Xinlong Wang, Wen Wang, Yue Cao, Chunhua Shen, and Tiejun Huang.
\newblock Images speak in images: A generalist painter for in-context visual learning.
\newblock In \emph{CVPR}, 2023{\natexlab{a}}.

\bibitem[Wang et~al.(2023{\natexlab{b}})Wang, Zhang, Cao, Wang, Shen, and Huang]{SegGPT}
Xinlong Wang, Xiaosong Zhang, Yue Cao, Wen Wang, Chunhua Shen, and Tiejun Huang.
\newblock {SegGPT:} segmenting everything in context.
\newblock In \emph{ICCV}, 2023{\natexlab{b}}.

\bibitem[Wang et~al.(2024)Wang, Yang, Chen, Wang, Guo, Chau, Liu, Qiao, Kot, and Wen]{wang2024sinsr}
Yufei Wang, Wenhan Yang, Xinyuan Chen, Yaohui Wang, Lanqing Guo, Lap-Pui Chau, Ziwei Liu, Yu~Qiao, Alex~C Kot, and Bihan Wen.
\newblock {SinSR}: diffusion-based image super-resolution in a single step.
\newblock In \emph{CVPR}, 2024.

\bibitem[Wang et~al.(2023{\natexlab{c}})Wang, Jiang, Lu, He, Chen, Wang, Zhou, et~al.]{wang2023context}
Zhendong Wang, Yifan Jiang, Yadong Lu, Pengcheng He, Weizhu Chen, Zhangyang Wang, Mingyuan Zhou, et~al.
\newblock In-context learning unlocked for diffusion models.
\newblock \emph{NeurIPS}, 2023{\natexlab{c}}.

\bibitem[Wei et~al.(2018)Wei, Wang, Yang, and Liu]{Chen2018Retinex}
Chen Wei, Wenjing Wang, Wenhan Yang, and Jiaying Liu.
\newblock Deep retinex decomposition for low-light enhancement.
\newblock In \emph{BMVC}, 2018.

\bibitem[Xia et~al.(2023)Xia, Zhang, Wang, Wang, Wu, Tian, Yang, and Van~Gool]{xia2023diffir}
Bin Xia, Yulun Zhang, Shiyin Wang, Yitong Wang, Xinglong Wu, Yapeng Tian, Wenming Yang, and Luc Van~Gool.
\newblock {DiffIR}: Efficient diffusion model for image restoration.
\newblock In \emph{ICCV}, 2023.

\bibitem[Xiao et~al.(2025)Xiao, Wang, Zhou, Yuan, Xing, Yan, Li, Wang, Huang, and Liu]{xiao2025omnigen}
Shitao Xiao, Yueze Wang, Junjie Zhou, Huaying Yuan, Xingrun Xing, Ruiran Yan, Chaofan Li, Shuting Wang, Tiejun Huang, and Zheng Liu.
\newblock Omnigen: Unified image generation.
\newblock In \emph{CVPR}, 2025.

\bibitem[Yang et~al.(2023)Yang, Peng, Shen, Yang, Hu, Qiu, Koike, et~al.]{yang2023imagebrush}
Yifan Yang, Houwen Peng, Yifei Shen, Yuqing Yang, Han Hu, Lili Qiu, Hideki Koike, et~al.
\newblock {ImageBrush:} learning visual in-context instructions for exemplar-based image manipulation.
\newblock \emph{NeurIPS}, 2023.

\bibitem[Zamir et~al.(2018)Zamir, Sax, , Shen, Guibas, Malik, and Savarese]{zamir2018taskonomy}
Amir~R Zamir, Alexander Sax, , William~B Shen, Leonidas Guibas, Jitendra Malik, and Silvio Savarese.
\newblock Taskonomy: Disentangling task transfer learning.
\newblock In \emph{CVPR}, 2018.

\bibitem[Zhang et~al.(2024{\natexlab{a}})Zhang, Jiang, Guo, Yan, Pan, Dong, Qiao, Gao, and Li]{zhangpersonalize}
Renrui Zhang, Zhengkai Jiang, Ziyu Guo, Shilin Yan, Junting Pan, Hao Dong, Yu~Qiao, Peng Gao, and Hongsheng Li.
\newblock Personalize segment anything model with one shot.
\newblock In \emph{ICLR}, 2024{\natexlab{a}}.

\bibitem[Zhang et~al.(2024{\natexlab{b}})Zhang, Doughty, and Snoek]{zhang2024low}
Yunhua Zhang, Hazel Doughty, and Cees~GM Snoek.
\newblock Low-resource vision challenges for foundation models.
\newblock In \emph{CVPR}, 2024{\natexlab{b}}.

\bibitem[Zhang et~al.(2025)Zhang, Yuan, Song, Wang, and Liu]{zhang2025easycontrol}
Yuxuan Zhang, Yirui Yuan, Yiren Song, Haofan Wang, and Jiaming Liu.
\newblock {EasyControl}: Adding efficient and flexible control for diffusion transformer.
\newblock In \emph{ICCV}, 2025.

\bibitem[Zhao et~al.(2024)Zhao, Fan, Kou, Qin, Gu, Wu, Xu, Zhu, Wang, and Gao]{zhao2024instructbrush}
Ruoyu Zhao, Qingnan Fan, Fei Kou, Shuai Qin, Hong Gu, Wei Wu, Pengcheng Xu, Mingrui Zhu, Nannan Wang, and Xinbo Gao.
\newblock {InstructBrush:} learning attention-based instruction optimization for image editing.
\newblock \emph{arXiv preprint arXiv:2403.18660}, 2024.

\end{thebibliography}
\bibliographystyle{iclr2026_conference}

\newpage
\appendix
\section*{Appendix}
This appendix provides supplementary information not included in the main paper due to space constraints. 
Specifically, it includes details of the Visual Relation Dataset (VisRel) (Section~\ref{visrel}), additional explanations of test-time scaling (Section~\ref{app:tts}), further ablation studies (Section~\ref{app:ablations}), extended analysis of test-time task generalization (Section~\ref{test-time}), and additional results (Section~\ref{app:moremoreresults}).

\section{Visual Relation Dataset (VisRel) Details}
\label{visrel}
The Visual Relation Dataset (VisRel) is a diverse collection of 2D visual tasks reformulated as visual transformations ($A \rightarrow A'$). It spans a wide range of task types and annotation formats, enabling the modeling of a unified visual relation space. It aims to trigger cross-task generalization and test-time adaptation via relation-space interpolation.
VisRel integrates heterogeneous datasets, each of which contributing different visual relations. For clarity, we categorize them into four groups based on their underlying task semantics: (1) Image Restoration and Enhancement,  (2) Physical and Geometric Perception, (3) Semantic Perception, and (4) Generative Manipulation.

Table~\ref{tab:dataset} provides a detailed overview of the datasets included in VisRel.  For each dataset, we list the task type, the visual transformation (input-output pair) that defines the task, and the annotation source. 
This diverse and well-structured dataset provides the foundation for our visual in-context learning framework, enabling PICO to generalize to novel user-personalized visual transformations at test time.  

\begin{table}[!ht]
\caption{\textbf{Summary of datasets in VisRel.}
Each dataset is represented by its task type, exemplar relation $(A \rightarrow A')$, and annotation source. \textbf{Ground Truth} denotes annotations provided by the original dataset, while \textbf{Human-labeled} indicates annotations created by us. }
\centering
\begin{adjustbox}{width=\linewidth}
\begin{tabular}{@{}l|l|l|l@{}}
\toprule
\textbf{Dataset} & \textbf{Task Type} & \textbf{Visual Relation ($A \rightarrow A'$)} & \textbf{Annotation Source} \\
\midrule
\multicolumn{4}{@{}c}{\textbf{Restoration /  Enhancement}} \\
DIV2K~\citep{Agustsson_2017_CVPR_Workshops} & Deblurring & Blurry Image $\rightarrow$ Clean Image & Ground Truth \\
DIV2K~\citep{Agustsson_2017_CVPR_Workshops} & Denoising & Noisy Image $\rightarrow$ Clean Image & Ground Truth \\
Synthetic Rain~\citep{Kui_2020_CVPR}  & Deraining & Rainy Image $\rightarrow$ Clean Image & Ground Truth \\
Dense-Haze \cite{Dense-Haze_2019}  & Dehazing & Hazy Image $\rightarrow$ Clean Image & Ground Truth \\
LOL~\citep{Chen2018Retinex}  & Low-Light Enhancement & 
Low-Light Image $\rightarrow$ Enhanced Image & Ground Truth \\
\midrule
\multicolumn{4}{@{}c}{\textbf{Physical / Geometric Perception}} \\
Taskonomy~\citep{zamir2018taskonomy} & Surface Normal Estimation  & RGB Image $\rightarrow$ Surface Normal Map  & Ground Truth \\
Taskonomy~\citep{zamir2018taskonomy} & Euclidean Distance Estimation & RGB Image $\rightarrow$ Distance Map & Ground Truth \\
Taskonomy~\citep{zamir2018taskonomy} & Z-buffer Depth Estimation & RGB Image $\rightarrow$ Z-buffer Map & Ground Truth \\
Taskonomy~\citep{zamir2018taskonomy}  & Principal Curvature Estimation & RGB Image $\rightarrow$ Curvature Map & Ground Truth \\
Taskonomy~\citep{zamir2018taskonomy}  & Reshading & RGB Image $\rightarrow$ Re-rendered Image & Ground Truth \\
Taskonomy~\citep{zamir2018taskonomy}  & 2D Keypoint Estimation & RGB Image $\rightarrow$ 2D Keypoint Heatmap & Ground Truth \\
Taskonomy~\citep{zamir2018taskonomy}  & 3D Keypoint Estimation & RGB Image $\rightarrow$ 3D Keypoint Heatmap & Ground Truth \\
Taskonomy~\citep{zamir2018taskonomy}  & Occlusion Edge Detection & RGB Image $\rightarrow$ Occlusion Edge Map & Ground Truth \\
Taskonomy~\citep{zamir2018taskonomy}  & Texture Edge Detection & RGB Image $\rightarrow$ Texture Edge Map & Ground Truth \\
\midrule
\multicolumn{4}{@{}c}{\textbf{Semantic Perception}} \\
MS-COCO~\citep{lin2014microsoft} & Instance Segmentation & Image $\rightarrow$ Instance Masks & Ground Truth \\
MS-COCO~\citep{lin2014microsoft} & Panoptic Segmentation & Image $\rightarrow$ Panoptic Masks & Ground Truth \\
MS-COCO~\citep{lin2014microsoft} & Semantic Segmentation & Image $\rightarrow$ Class Masks  & Ground Truth \\
DIS5K~\citep{qin2022} & Dichotomous Segmentation & Image $\rightarrow$ Binary Mask & Ground Truth \\
CORe50~\citep{lomonaco2017core50} & Object Detection & Image $\rightarrow$ Bounding Boxes & Human-labeled \\
MS-COCO~\citep{lin2014microsoft} & Human Pose Estimation & Image $\rightarrow$ Keypoint Map & Ground Truth \\
OPRA~\citep{demo2vec2018cvpr} & Accordance Detection & Image $\rightarrow$ Highlighted Accordance Part & Ground Truth \\
\midrule
\multicolumn{4}{@{}c}{\textbf{Generative Manipulation}} \\
DIV2K~\citep{Agustsson_2017_CVPR_Workshops} & Inpainting & Masked Image $\rightarrow$ Completed Image & Ground Truth \\
MS-COCO~\citep{lin2014microsoft} & Style Transfer & Image $\rightarrow$ Stylized Image & \citet{chung2024style} \\
PhotoDoodle~\citep{huang2025photodoodle} & Doodling & Image $\rightarrow$ Image with Doodles  & Ground Truth \\
Cat~\citep{crawfordcats} & Sticker Addition & Image $\rightarrow$ Image with Stickers & Human-labeled \\
PaintBucket~\citep{InclusionMatching2024} & Line Art Colorization & Line Art $\rightarrow$ Colored Image & Ground Truth \\
ReNé~\citep{Toschi_2023_CVPR} & Object Relighting & Image under Light A $\rightarrow$ Light A'
  & Ground Truth \\
\bottomrule
\end{tabular}
\label{tab:dataset}
\end{adjustbox}
\end{table}


\section{Additional Details about Test-time scaling}
\label{app:tts}

\subsection{Attention–Guided Seed Selection Procedure}
\label{app:tts:defs}

Algorithm~\ref{alg:seed-select} details the test-time seed selection procedure.

\begin{algorithm}[t]
\caption{Attention–Guided Seed Selection (AGSS)}
\label{alg:seed-select}
\begin{algorithmic}[1]
\Require quad-grid $I$ with BR placeholder X, seeds $\mathcal{S}$, warmup steps $\mathcal{I}_{\mathrm{probe}}{=}\{0,\dots,i_{\mathrm{warm}}\}$, critical blocks $\mathcal{B}^\dagger$
\State $c_{\text{vp}}\gets \mathcal{E}([A,A',B])$
\ForAll{$s\in\mathcal{S}$} \Comment{batched warmup}
  \State $x^{(s)}_{0}\sim\mathcal{N}(0,I)$
  \For{$i\in \mathcal{I}_{\mathrm{probe}}$}
    \State advance one solver step on BR only; record $W^{(b,h)}_{s,i}$ for $b\in\mathcal{B}^{\dagger}$
  \EndFor
  \State compute $\{p^{\text{br}}_{s,b,i},p^{\text{vp}}_{s,b,i}\}$, then $D_{\text{br}}(s),D_{\text{vp}}(s)$
\EndFor
\State $S_{\text{pivot}}(s)\leftarrow z(D_{\text{br}}(s))-z(D_{\text{vp}}(s))$ \quad(for all $s$, per-image $z$-norm)
\State $s^\star\leftarrow \arg\max_{s\in\mathcal{S}} S_{\text{pivot}}(s)$
\State continue denoising from the cached BR state of $s^\star$ to obtain $\hat x_0$; output $B'=\mathcal{D}(\hat x_0)$
\end{algorithmic}
\end{algorithm}

\subsection{Statistical Analysis}
\label{sec:seed-scorer}

\noindent{\textbf{Notation.}}
Let $s\in\mathcal{S}$ be a seed, $b$ a cross-attention block, $h\in\{1,\dots,H\}$ a head, and $i$ an early step in a warmup window $\mathcal{I}_{\mathrm{probe}}=\{0,\dots,i_{\mathrm{warm}}\}$.
Denote the attention by $W^{(b,h)}_{s,i}\!\in\!\mathbb{R}^{Q\times K}$.
Queries are BR (target) tokens $\mathcal{Q}_{\text{br}}$; keys are partitioned into disjoint sets:
$\mathcal{K}_{\text{br}}$ (BR), $\mathcal{K}_{\text{vp}}$ (context \{A,A',B\}), and $\mathcal{K}_{\text{txt}}$ (text).
Average masses from BR queries at step $i$ are
\begin{equation}
\label{eq:attention-mass}
p^{\star}_{s,b,i}
=\frac{1}{H\,|\mathcal{Q}_{\text{br}}|}\!
\sum_{h=1}^H \sum_{q\in\mathcal{Q}_{\text{br}}} \sum_{k\in\mathcal{K}_{\star}}
W^{(b,h)}_{s,i}[q,k],\qquad \star\in\{\text{br},\text{vp},\text{txt}\}.
\end{equation}

\noindent{\textbf{Setting.}}
We evaluate $100$ images $\times 16$ seeds with BR-only warmup on steps $\{0,1,2\}$.
For each (image, seed) we compute atoms over $\mathcal{B}^{\dagger}$ and $z$-normalize across the $16$ seeds (per image).
Performance is mIoU on the BR quadrant (bIoU yields the same ordering).

With $i^\ast{=}2$,
\begin{align}
L_{\text{br}}(s)&=\tfrac{1}{|\mathcal{B}^{\dagger}|}\textstyle\sum_{b\in\mathcal{B}^{\dagger}} p^{\text{br}}_{s,b,i^\ast}, &
D_{\text{br}}(s)&=\tfrac{1}{|\mathcal{B}^{\dagger}|}\textstyle\sum_{b\in\mathcal{B}^{\dagger}}\big(p^{\text{br}}_{s,b,i^\ast}-p^{\text{br}}_{s,b,0}\big),\\
L_{\text{vp}}(s)&=\tfrac{1}{|\mathcal{B}^{\dagger}|}\textstyle\sum_{b\in\mathcal{B}^{\dagger}} p^{\text{vp}}_{s,b,i^\ast}, &
D_{\text{vp}}(s)&=\tfrac{1}{|\mathcal{B}^{\dagger}|}\textstyle\sum_{b\in\mathcal{B}^{\dagger}}\big(p^{\text{vp}}_{s,b,i^\ast}-p^{\text{vp}}_{s,b,0}\big),\\
L_{\text{txt}}(s)&=\tfrac{1}{|\mathcal{B}^{\dagger}|}\textstyle\sum_{b\in\mathcal{B}^{\dagger}} p^{\text{txt}}_{s,b,i^\ast}, &
D_{\text{txt}}(s)&=\tfrac{1}{|\mathcal{B}^{\dagger}|}\textstyle\sum_{b\in\mathcal{B}^{\dagger}}\big(p^{\text{txt}}_{s,b,i^\ast}-p^{\text{txt}}_{s,b,0}\big).
\end{align}
$L_{\bullet}$ capture \emph{where} attention lands by the end of warmup; $D_{\bullet}$ capture \emph{how} it pivots.

\begin{figure}[t]
  \centering

  \begin{minipage}[t]{0.58\linewidth}
    \centering
    \includegraphics[height=4.2cm]{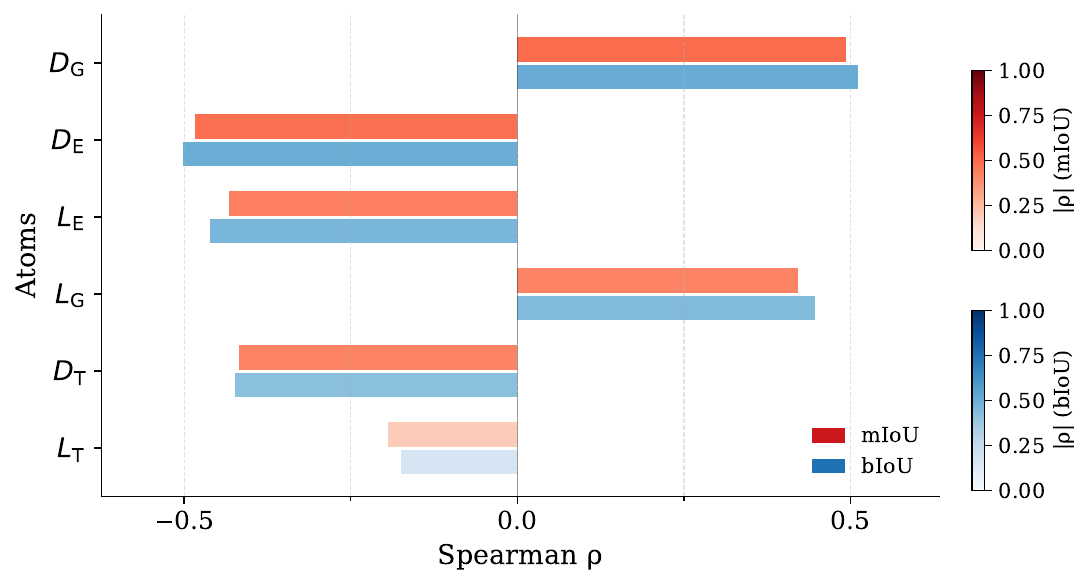}
    \captionof{figure}{Spearman $\rho$ between $z$-normalized attention atoms and mIoU/bIoU across seeds.}
    \label{fig:spearman}
  \end{minipage}
  \hfill
  \begin{minipage}[t]{0.4\linewidth}
    \centering
    \includegraphics[height=4.2cm]{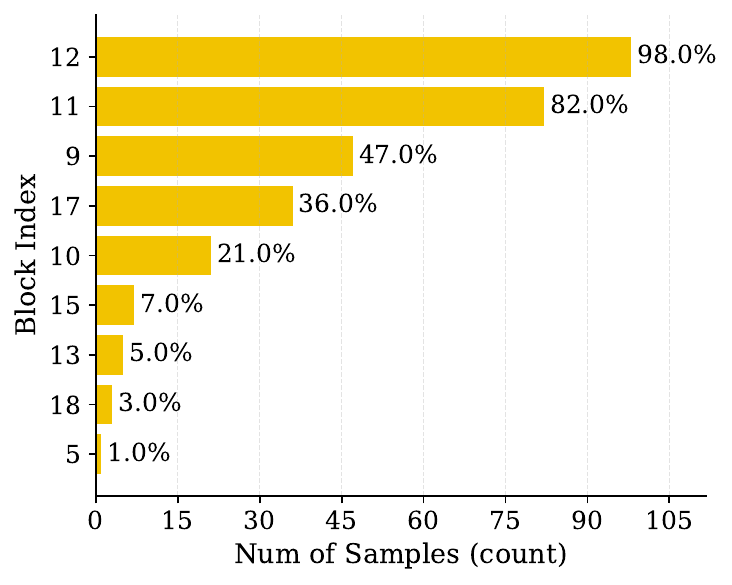}
    \captionof{figure}{Top-10 most seed-sensitive cross-attention blocks in \citet{flux2024}.}
    \label{fig:blocks}
  \end{minipage}

\end{figure}

\paragraph{Correlation map.}
We compute Spearman rank correlations between $z$-normalized atoms and mIoU across seeds (pooled over images). As shown in Figure~\ref{fig:spearman}, the results align with our policy: BR (target) atoms are positively correlated with mIoU, visual-context atoms (from \{A,A',B\}) are negatively correlated, and text atoms exhibit weak correlations.
%

\subsubsection{Identifying Critical Blocks}
\label{app:tts:stats:crit}
Seed sensitivity concentrates in a narrow stage.
For each image and block $b$, we define a simple visual context–target gap
\begin{equation}
\label{eq:attention-gap}
\mathrm{gap}^{(s)}_{b,i}=p^{\text{vp}}_{s,b,i}-p^{\text{br}}_{s,b,i},\quad i\in\{0,1,2\}.
\end{equation}
We scan cross-attention blocks and summarize each block by two statistics over the warmup: (i) a level
term (the average context–target gap), and (ii) a growth term (the change in the gap from the first
to the last warmup step).
Blocks whose level/growth exhibit high variance across seeds are most discriminative for seed ranking.
We therefore retain the top few blocks as the critical set $\mathcal{B}^{\dagger}$ (we use $K{=}3$).
Aggregated across images, the same small stage (e.g., blocks $\{9,11,12\}$) consistently emerges (see Figure~\ref{fig:blocks}).

\section{Additional Studies}
\label{app:ablations}
\subsection{Effect of Visual Prompts}
\label{abl_img}

In Section~\ref{main:freeform-seg}, we demonstrate that PICO supports free-form inputs for personalized segmentation tasks. Here, we further investigate the role of visual prompts—\textit{i.e.}, in-context input-output exemplars ($A \rightarrow A'$)—in providing fine-grained control over output behavior across additional representative task categories.
Given a fixed query image $B$ and text prompt, PICO flexibly adapts to different task intents by interpreting the visual demonstration $(A \rightarrow A')$, producing diverse and context-appropriate outputs $B'$.
%
%
\begin{figure}[!b]
  \centering
  \includegraphics[width=\linewidth]{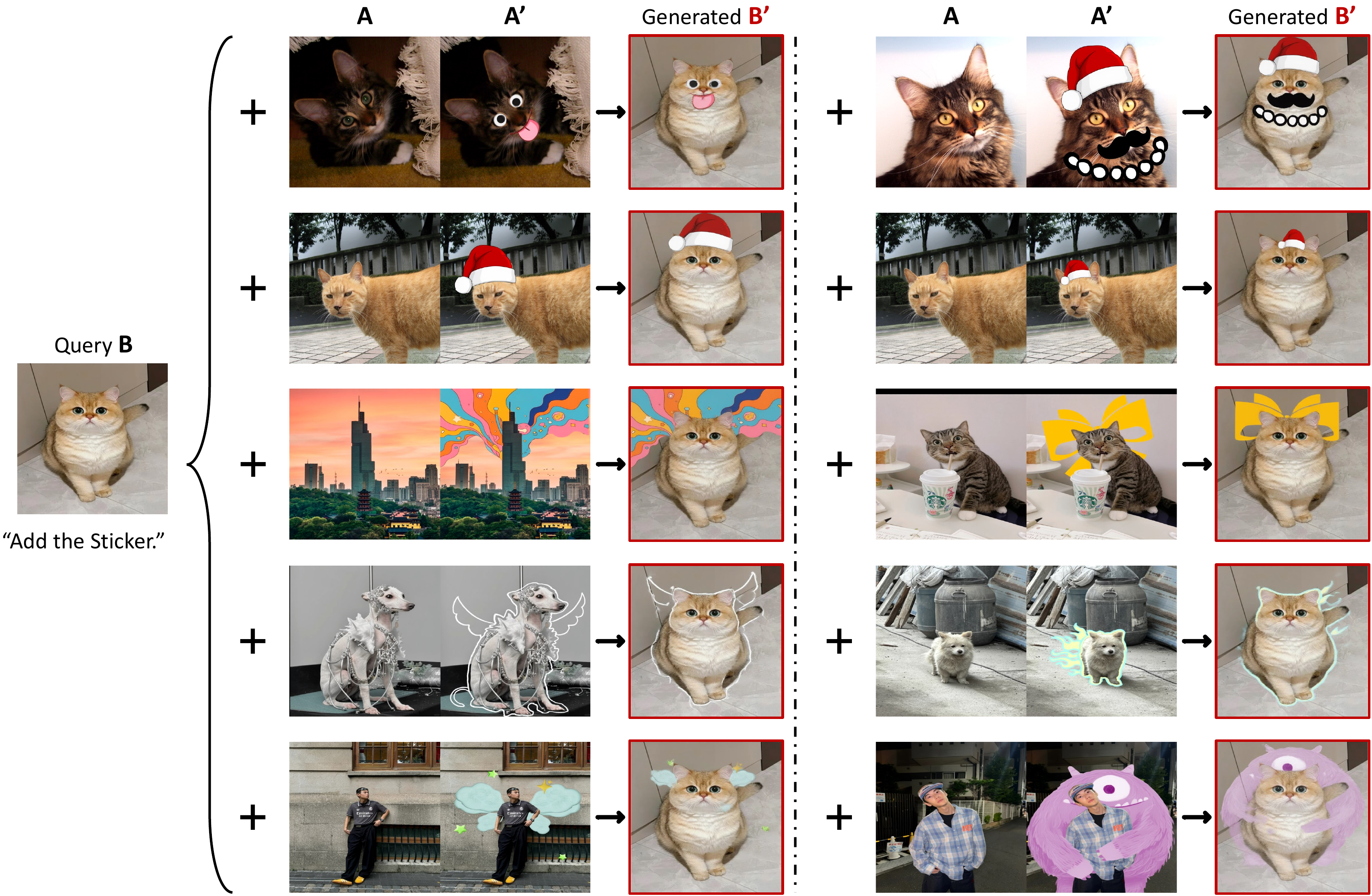}
  \caption{\textbf{Context-aware sticker addition with visual prompt control.} 
Given the same query image $B$ and text prompt (``Add the sticker''), PICO generates diverse outputs $B'$ solely based on visual prompt $(A \rightarrow A')$. 
The model captures variation in object type, position, and scale, demonstrating precise spatial and semantic interpretation from visual prompts.}
  \label{fig:image_sticker}
\end{figure}

\textbf{Context-Aware Sticker Addition.}  
Figure~\ref{fig:image_sticker} shows how the visual prompt controls what customized object or doodling is added, where it is placed, and how it is scaled. For example, the size of a Christmas hat (Row 2) changes based on the visual exemplar, despite the same text prompt (``Add the sticker”).
This task highlights the limitations of text-only instructions and the strength of visual exemplars for conveying spatial and compositional intent.

\begin{figure}[ht]
  \centering
  \includegraphics[width=\linewidth]{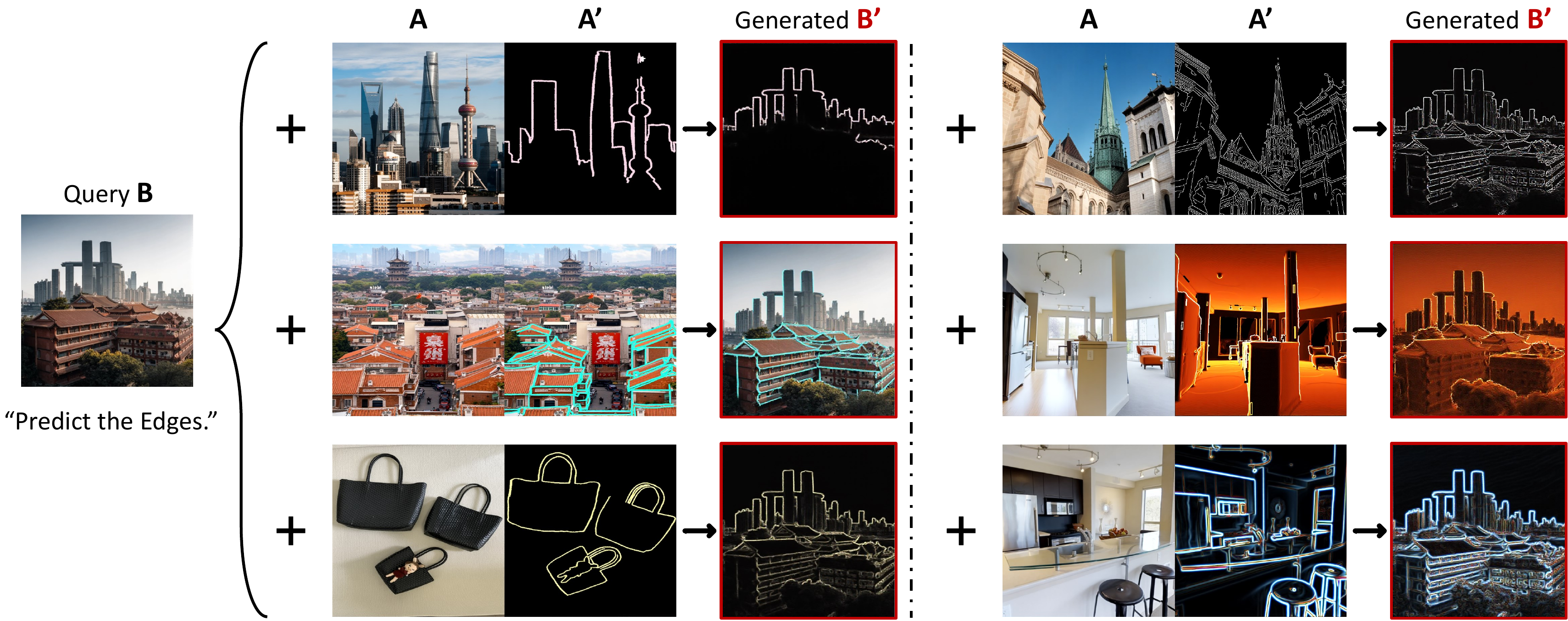}
  \caption{\textbf{Personalized edge detection with visual prompt control.} 
    Given the same query image $B$ and text prompt (``Predict the edges''), PICO generates diverse outputs $B'$ solely based on visual exemplars $(A \rightarrow A')$. 
    The model adapts spatial focus (\textit{e.g.}, top or bottom) and edge style (\textit{e.g.}, canny, euclidean, texture), guided entirely by visual cues.}
  \label{fig:image_edge}
\end{figure}

\textbf{Personalized Edge Detection.}  
As shown in Figure~\ref{fig:image_edge}, PICO handles edge detection tasks defined by spatial constraints and style cues in the visual prompts. The model is able to adaptively predict edges of specific regions (\textit{e.g.}, top vs. bottom) or emulate particular edge styles (\textit{e.g.}, Canny vs. Euclidean vs. texture-based) without making any changes to the text prompt (``Predict the edges'').

These results confirm that PICO effectively comprehends the visual relation conveyed by the in-context input-output pairs, and applies the underlying visual logic to query images.
The quad-grid in-context format provides a strong structural prior for visual reasoning, enabling flexible, controllable, and free-form at test time.

\begin{table}[!h]
\caption{\textbf{Ablation on task-type effects for personalized segmentation.} 
The best results are highlighted in \textbf{bold}, and the second-best are \underline{underlined}.}
\vspace{-2mm}
\centering
\begin{adjustbox}{width=\linewidth}
\begin{tabular}{@{}l|lll|lll|lll|lll@{}}
\toprule
\multirow{2}{*}{Method} & \multicolumn{3}{c|}{PerSeg} & \multicolumn{3}{c|}{DOGS} & \multicolumn{3}{c|}{PODS} & \multicolumn{3}{c}{PerMIS} \\
 & mIoU$\uparrow$ & bIoU$\uparrow$ & F1$\uparrow$ & mIoU$\uparrow$ & bIoU$\uparrow$ & F1$\uparrow$ & mIoU$\uparrow$ & bIoU$\uparrow$ & F1$\uparrow$ & mIoU$\uparrow$ & bIoU$\uparrow$ & F1$\uparrow$ \\
\midrule
\rowcolor{gray!20}
PICO              & \underline{90.97} & 76.13 & 62.82 & 71.02 & 54.71 & 49.84 & \textbf{68.72} & \textbf{60.26} & \underline{44.88} & \underline{49.52} & \textbf{33.63} & 14.90 \\

Remove physical   & 82.68~\textcolor{red}{$\downarrow$}  & 72.45~\textcolor{red}{$\downarrow$} & 54.11~\textcolor{red}{$\downarrow$} & 62.47~\textcolor{red}{$\downarrow$} & 52.56~\textcolor{red}{$\downarrow$} & 51.36~\textcolor{green}{$\uparrow$} & 50.44~\textcolor{red}{$\downarrow$} & 50.97~\textcolor{red}{$\downarrow$} & 35.96~\textcolor{red}{$\downarrow$} & 42.12~\textcolor{red}{$\downarrow$} & 27.72~\textcolor{red}{$\downarrow$} & 12.02~\textcolor{red}{$\downarrow$} \\

Remove low-level  & 89.71~\textcolor{red}{$\downarrow$} & \underline{76.14}~\textcolor{green}{$\uparrow$} & \underline{62.89}~\textcolor{green}{$\uparrow$} & \textbf{75.52}~\textcolor{green}{$\uparrow$} & \textbf{60.86}~\textcolor{green}{$\uparrow$} & \underline{60.47}~\textcolor{green}{$\uparrow$} & 61.29~\textcolor{red}{$\downarrow$} & \underline{59.52}~\textcolor{red}{$\downarrow$} & \textbf{45.91}~\textcolor{green}{$\uparrow$} & 48.32~\textcolor{red}{$\downarrow$} & \underline{32.60}~\textcolor{red}{$\downarrow$} & \underline{15.22}~\textcolor{green}{$\uparrow$} \\

Remove Generative & 70.88~\textcolor{red}{$\downarrow$} & 59.68~\textcolor{red}{$\downarrow$} & 39.15~\textcolor{red}{$\downarrow$} & 63.02~\textcolor{red}{$\downarrow$} & 52.11~\textcolor{red}{$\downarrow$} & 44.26~\textcolor{red}{$\downarrow$} & 22.86~\textcolor{red}{$\downarrow$} & 23.29~\textcolor{red}{$\downarrow$} & 11.38~\textcolor{red}{$\downarrow$} & 35.07~\textcolor{red}{$\downarrow$} & 21.56~\textcolor{red}{$\downarrow$} & 7.37~\textcolor{red}{$\downarrow$} \\

Remove Semantic   & \textbf{92.90}~\textcolor{green}{$\uparrow$}  & \textbf{78.11}~\textcolor{green}{$\uparrow$} & \textbf{63.98}~\textcolor{green}{$\uparrow$} & \underline{74.93}~\textcolor{green}{$\uparrow$} & \underline{58.54}~\textcolor{green}{$\uparrow$} & \textbf{61.19}~\textcolor{green}{$\uparrow$} & \underline{62.07}~\textcolor{red}{$\downarrow$} & 57.64~\textcolor{red}{$\downarrow$} & 44.41~\textcolor{red}{$\downarrow$} & \textbf{50.58}~\textcolor{green}{$\uparrow$} & 32.58~\textcolor{red}{$\downarrow$} & \textbf{16.53}~\textcolor{green}{$\uparrow$} \\
\bottomrule
\end{tabular}
\end{adjustbox}
\label{tab:ablation_components}
\end{table}

\subsection{Task Type Ablation}
\label{abl_task}

Our VisRel dataset balances diversity across task types to holistically support personalization. We assess each family’s contribution by removing three tasks per family (30 samples) and retraining under identical settings.
Results in Table~\ref{tab:ablation_components} show:

\begin{itemize}
    \item \textbf{Physical/geometric tasks} (\textit{e.g.}, depth, reshading, 3D keypoints) consistently help. Removal reduces performance. 3D/spatial reasoning improves object boundary awareness.
    \item \textbf{Generative tasks} (\textit{e.g.}, doodling, relighting, line-art colorization) are critical. Removal causes big performance drop. These high-semantic/local tasks teach object-aligned editing essential for segmenting user-specific objects.
    \item \textbf{Low-level tasks} (\textit{e.g.}, deblurring, dehazing, low-light enhancement) are neutral but still valuable. Removal shows small changes. While not directly beneficial, they don’t harm performance, validating our inclusive design.
    \item \textbf{Semantic perception tasks} (\textit{e.g.}, stuff segmentation, object detection, affordance) can be conflicting. Interestingly, removal improves results. We hypothesize that class‑level labels may suppress fine instance‑level distinctions, which are essential for personalized segmentation.
\end{itemize}

\subsection{Spatial Position of the Grid Format}
\label{abl_grid}
During training and inference, we fixed the placeholder position (i.e., the cell of $B'$) in the $2\times2$ grid.  To evaluate the impact of placeholder positioning, we conducted the experiment by changing the curernt horizontal layout (\emph{Top–Bottom}, TB) to a vertical arrangement (\emph{Left–Right}, LR), using identical datasets.  

The results in Tables~\ref{tab:pos_personalized_seg}, \ref{tab:pos_personalized_generalization} show that grid positioning has minimal impact on overall model performance. In some cases, the LR layout improves personalized segmentation metrics. The grid layout (or, essentially, the positional embedding) has little impact, but the learned visual-relation space drives performance.

\begin{table}[!h]
\caption{\textbf{Ablation on grid layout for personalized image segmentation.} }
\vspace{-2mm}
\centering
\begin{adjustbox}{width=\linewidth}
\begin{tabular}{@{}l|ccc|ccc|ccc|ccc@{}}
\toprule
\multirow{2}{*}{Method} & \multicolumn{3}{c|}{PerSeg} & \multicolumn{3}{c|}{DOGS} & \multicolumn{3}{c|}{PODS} & \multicolumn{3}{c}{PerMIS} \\
 & mIoU$\uparrow$ & bIoU$\uparrow$ & F1$\uparrow$ & mIoU$\uparrow$ & bIoU$\uparrow$ & F1$\uparrow$ & mIoU$\uparrow$ & bIoU$\uparrow$ & F1$\uparrow$ & mIoU$\uparrow$ & bIoU$\uparrow$ & F1$\uparrow$ \\
\midrule
PICO (TB) & 90.97 & \textbf{76.13} & 62.82 & 71.02 & 54.71 & 49.84 & 68.72 & 60.26 & 44.88 & \textbf{49.52} & 33.63 & 14.90 \\
PICO (LR) & \textbf{91.73} & 75.87 & \textbf{63.53} & \textbf{75.90} & \textbf{58.44} & \textbf{58.71} & \textbf{70.27} & \textbf{61.79} & \textbf{45.88} & 47.69 & \textbf{33.64} & \textbf{17.34} \\
\bottomrule
\end{tabular}
\end{adjustbox}
\label{tab:pos_personalized_seg}
\end{table}

\begin{table}[!h]
\caption{\textbf{Ablation on grid layout for personalized test-time task generalization.} }
\vspace{-2mm}
\centering
\begin{adjustbox}{width=0.8\linewidth}
\begin{tabular}{@{}l|cc|cccc@{}}
\toprule
& \multicolumn{2}{c|}{(a) \textit{Deraining with Inpainting}} & \multicolumn{4}{c}{(b)\textit{ Inpainting with Stylization}} \\
Method & PSNR$\uparrow$ & SSIM$\uparrow$ & Gram$\downarrow$ & FID$\downarrow$ & LPIPS$\downarrow$ & ArtFID$\downarrow$ \\
\midrule
Ref        & $\infty$ & 1.00 & 17.29 & 1.71 & 0.62 & 4.38 \\
PICO (TB)  & 22.24 & \textbf{0.67} & \textbf{21.27} & 1.87 & 0.52 & \textbf{4.38} \\
PICO (LR)  & \textbf{22.42} & \textbf{0.67} & 21.51 & 1.87 & \textbf{0.51} & 4.39 \\
\bottomrule
\end{tabular}
\end{adjustbox}
\label{tab:pos_personalized_generalization}
\end{table}


\section{More on Test-time Tasks Generalization}
\label{test-time}

\subsection{Quantitative Evaluation}
\label{app:quantitative-details}
To complement the main results in Section~\ref{main:test-time}, we provide full details of the quantitative setups.  
We evaluate two representative composite tasks:  

\textbf{(1) Deraining with inpainting.}  
We evaluate $200$ images corrupted by rain and occlusions. We use:
(i) PSNR to assess pixel-level reconstruction fidelity, and 
(ii) SSIM to measure structural similarity between the predicted output $B'$ and the clean reference image $\text{Cleaned}(B)$.

\textbf{(2) Inpainting with Stylization.}  
We evaluate 265 stylized images across 40 style different styles, each stylized using StyleID~\citep{chung2024style} and then corrupted by watermarks or inpainting masks.
Evaluation metrics include:
(i) Gram Matrix Distance between $B'$ and the reference style image $A'$ to measure style fidelity, (ii) LPIPS between $B'$ and the original $\text{Cleaned}(B)$, to evaluate content preservation and occlusion removal, and (iii) ArtFID~\citep{chung2024style}, defined as $(\text{LPIPS} + 1)\cdot(\text{FID} + 1)$, which captures the overall trade-off between perceptual faithfulness and style fidelity.
As a reference upper bound, we include the ``ground truth'' result: applying StyleID~\citep{chung2024style} directly to the clean image $\text{Cleaned}(B)$ using the same target style as $A'$.

\subsection{Additional Quantitative comparisons}
\label{app:qualitative-comp}
\begin{figure}[!h]
  \centering
  \includegraphics[width=\linewidth]{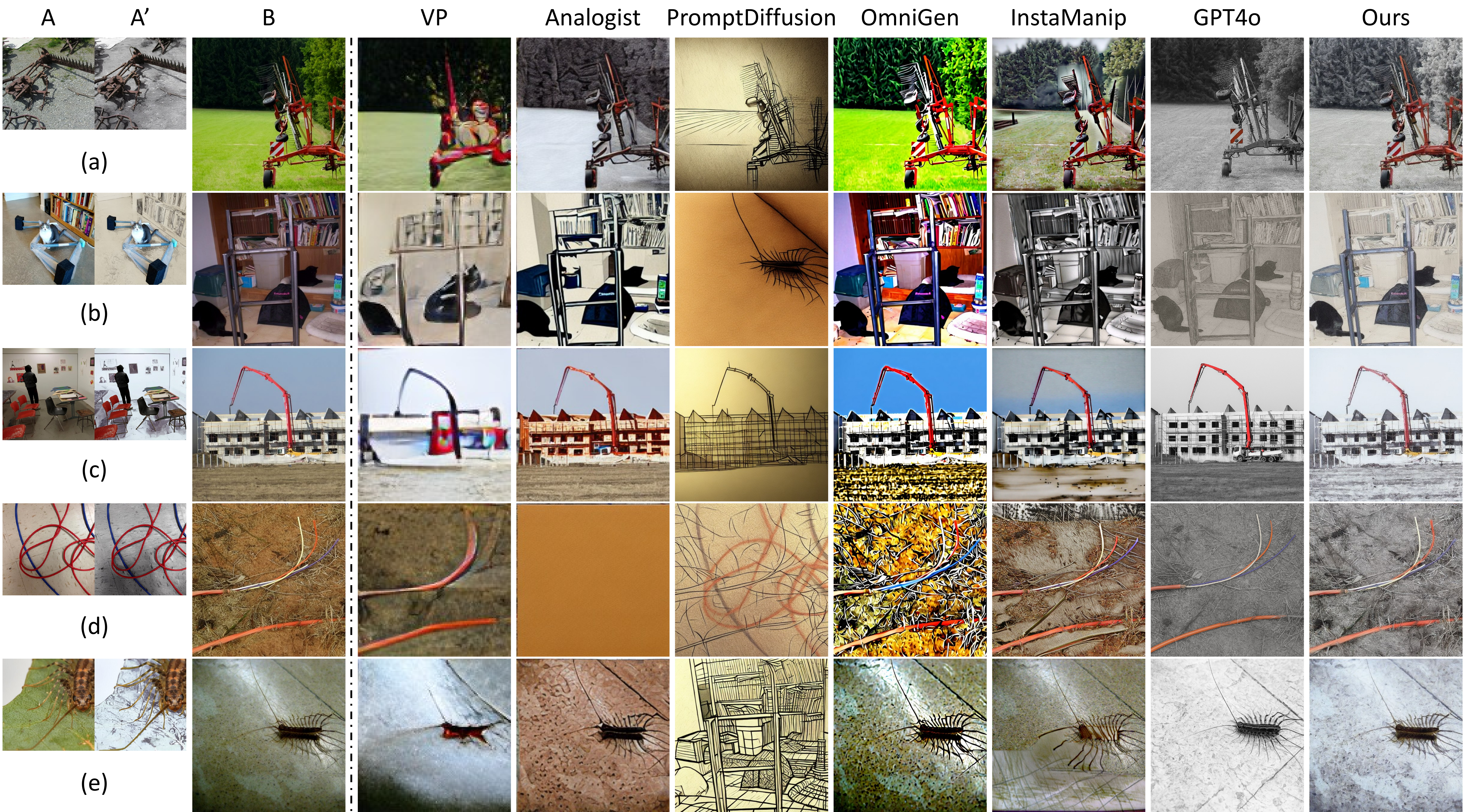}
  \caption{\textbf{Qualitative comparisons on background-only stylization.} PICO selectively stylizes the background while preserving the foreground.}
  \label{fig:bgstyle}
\end{figure}

\begin{figure}[t]
  \centering
  \includegraphics[width=\linewidth]{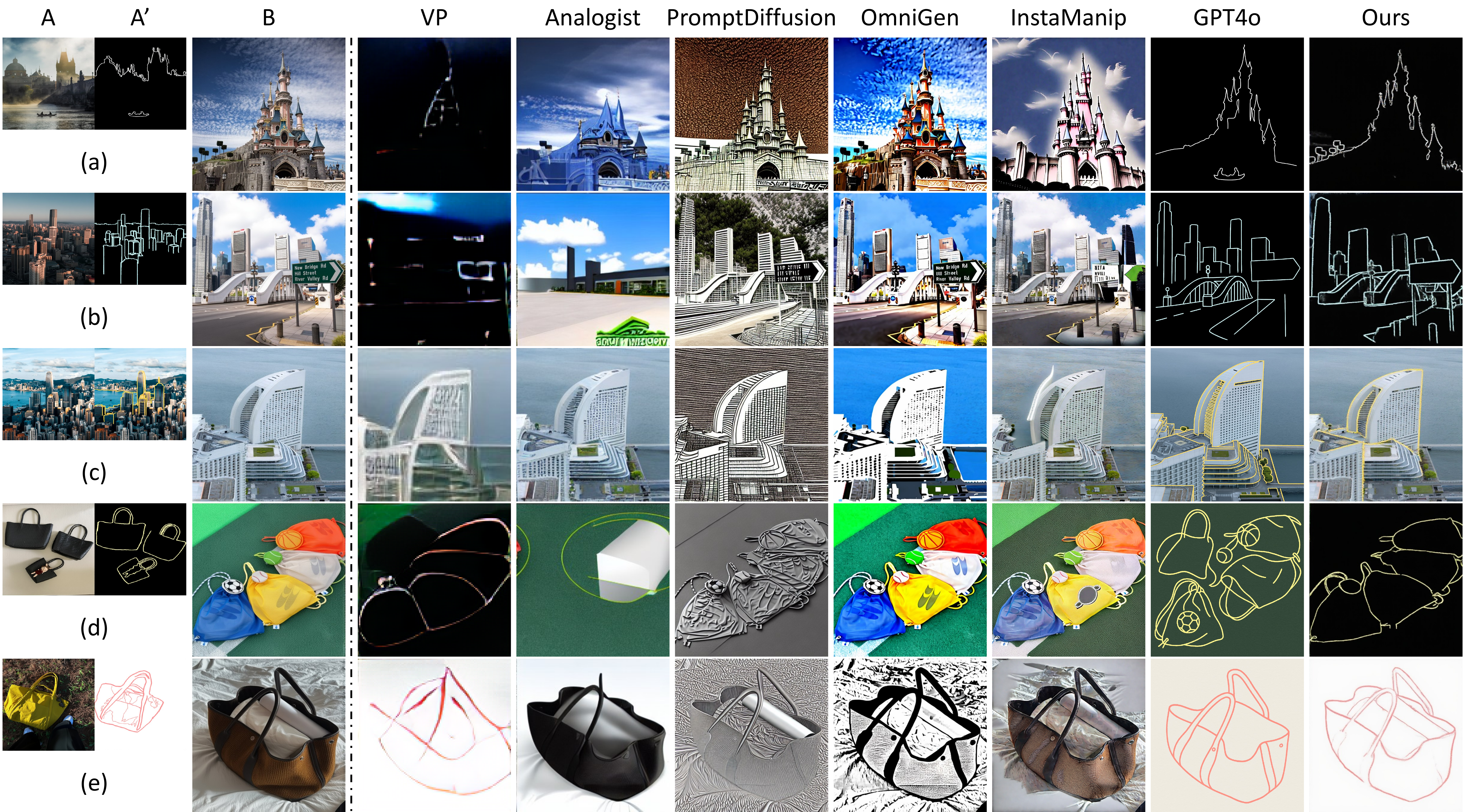}
  \caption{\textbf{Qualitative comparisons on edge detection with spatial constraints.} PICO accurately predicts personalized edge maps guided by the visual prompt.}
  \label{fig:edge}
\end{figure}

\begin{figure}[t]
  \centering
  \includegraphics[width=\linewidth]{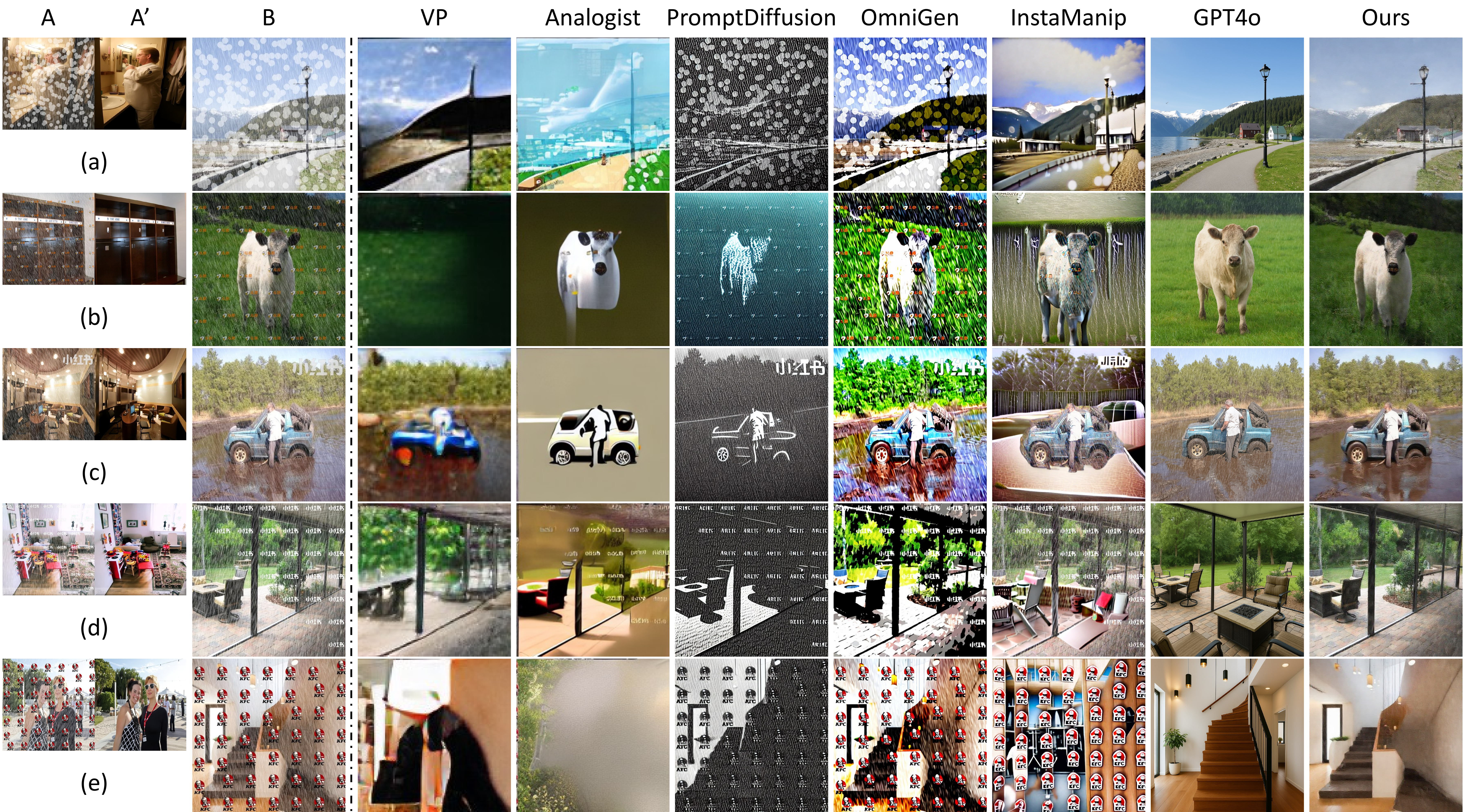}
  \caption{\textbf{Qualitative comparisons on joint deraining with inpainting.} PICO removes both rain and  occlusions simultaneously.}
  \label{fig:inpaintderain}
\end{figure}

\begin{figure}[t]
  \centering
  \includegraphics[width=\linewidth]{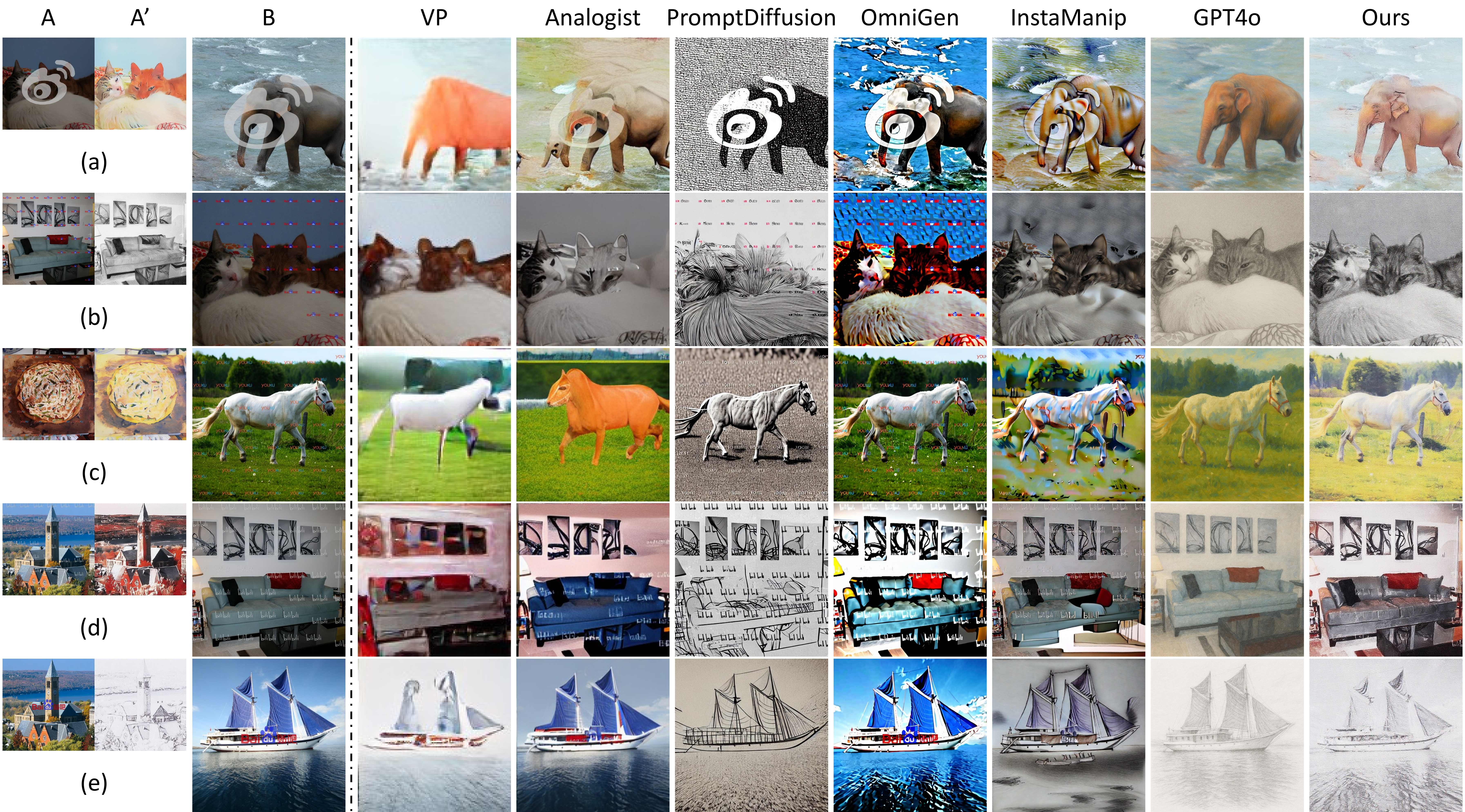}
  \caption{\textbf{Qualitative comparisons on watermark removal with stylization.} PICO removes occlusions while transferring target style.}
  \label{fig:inpaintstyle}
\end{figure}

\begin{figure}[t]
  \centering
  \includegraphics[width=\linewidth]{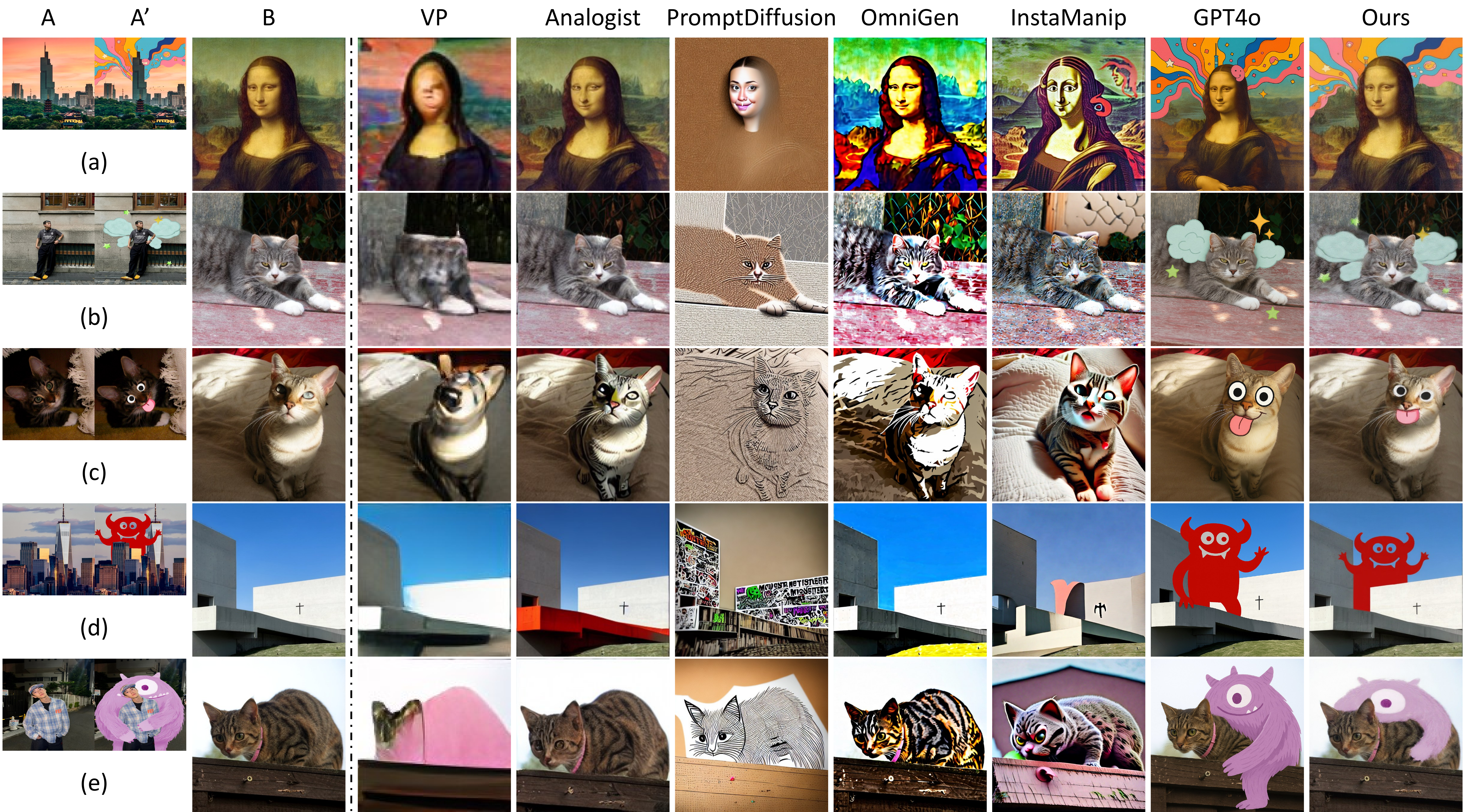}
  \caption{\textbf{Qualitative comparisons on context-aware sticker addition.} PICO learns from the visual exemplar where and how to place the sticker (\textit{e.g.}, object type, size, position).}
  \label{fig:sticker}
\end{figure}

Figures~\ref{fig:bgstyle}, \ref{fig:edge}, \ref{fig:inpaintderain}, \ref{fig:inpaintstyle}, \ref{fig:sticker} present additional qualitative comparisons across diverse test-time personalized tasks: background-only stylization, edge detection with spatial constraints, joint deraining with inpainting, watermark removal with stylization, and context-aware sticker addition.
PICO demonstrates consistent superiority in aligning with the task intent, as defined by in-context visual exemplar pair $(A \rightarrow A')$. GPT-4o shows strong semantic-level understanding but lacks precision in content fidelity and spatial alignment, especially in tasks that require geometric fidelity or pixel-aligned outputs.

\section{Additional Results}
\label{app:moremoreresults}

\begin{figure}[t]
  \centering
  \includegraphics[width=\linewidth]{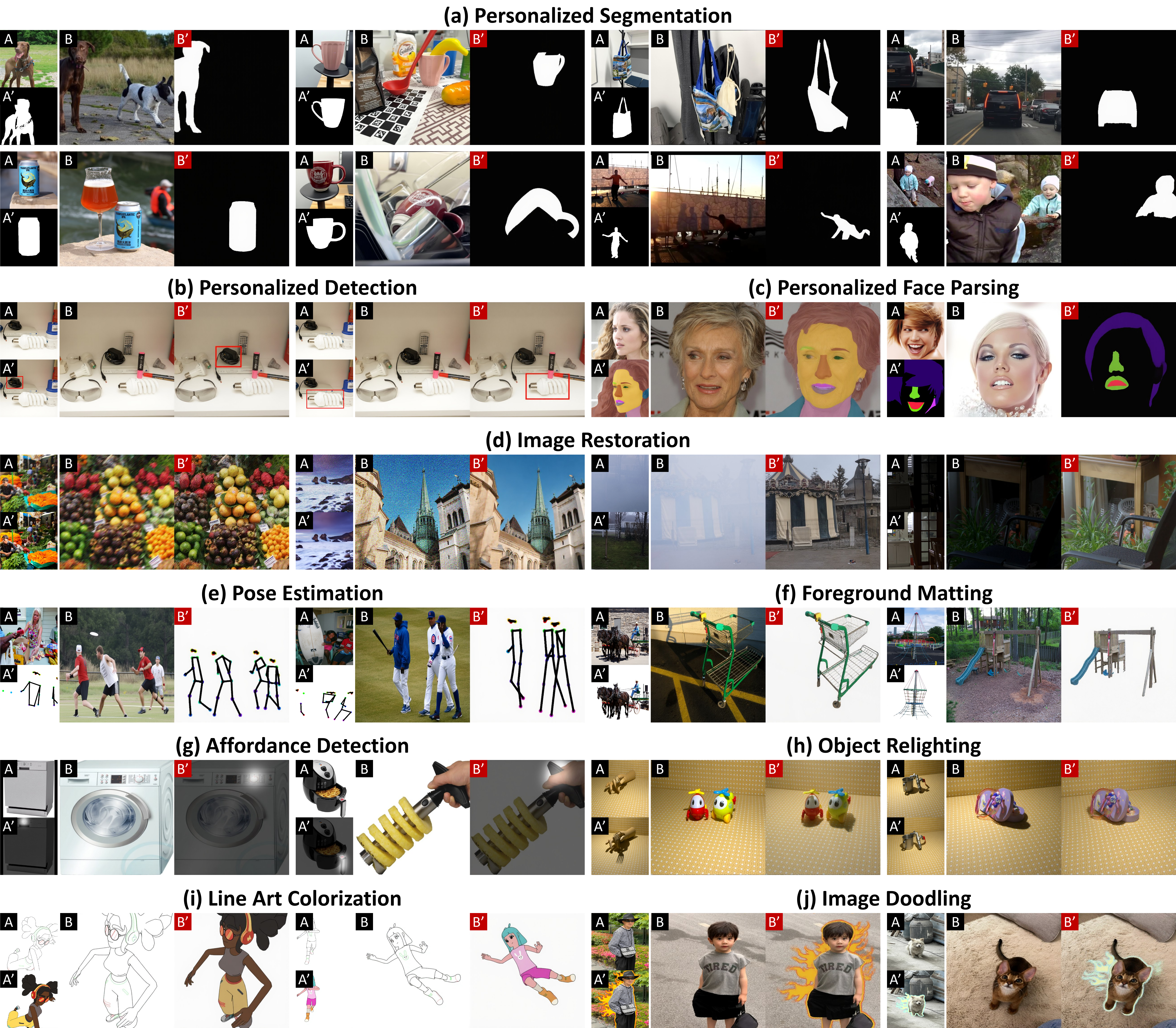}
  \caption{\textbf{Additional results generated by PICO.}}
  \label{fig:app:more-results}
\end{figure}

We present additional results generated by PICO across diverse tasks in Figure~\ref{fig:app:more-results}.
For personalized face parsing (Figure~\ref{fig:app:more-results}(c)), PICO leverages contextual appearance cues to consistently segment semantically identical components.  
Despite never being trained on facial data, the model performs well on this out-of-domain setting, demonstrating robustness and flexibility.  
PICO also supports a broad range of standard visual tasks spanning restoration, perception, and generation, as illustrated in Figure~\ref{fig:app:more-results}(d–k).  
While trained on these tasks, PICO generalizes effectively to novel object instances and scenarios with as few as $10$ example pairs per task.  
Notably, for object relighting, \textit{i.e.}, transforming an object under one lighting condition into another, PICO predicts physically plausible shadows aligned with previously unseen query objects (Figure~\ref{fig:app:more-results}(f)).  
These results suggest an implicit understanding of lighting and object interactions.
%




\end{document}